\documentclass[12pt, letterpaper]{report}   

\usepackage[bf]{caption} 
\setcaptionmargin{0.5in}
\usepackage{bu_math_thesis}
\usepackage{amsfonts}
\usepackage{amsmath}
\DeclareMathOperator*{\argmax}{arg\,max}

\graphicspath{{figures/}{tables/}}

\begin{document}
\setcitestyle{square}


\title{Psychophysical Evaluation of Deep Re-Identification Models}
\author{Hamish Nicholson}

\degree=2

\department{Department of Computer Science}
\university{Harvard University}
\faculty{School of Engineering and Applied Science}

\defenseyear{2020}
\degreeyear{2020}

\reader{First}{George Alvarez, PhD}{Professor, Psychology}
\reader{Second}{Stratos Idreos, PhD}{Associate Professor, Computer Science}
\reader{Third}{Hanspeter Pfister, PhD}{An Wang Professor,  Computer Science}

\majorprof{Primary Advisor}{MD MPH PhD}


\allowdisplaybreaks




\maketitle




\newpage
\addcontentsline{toc}{chapter}{Acknowledgements}
\section*{Acknowledgments}
I would like to express my appreciation to Professor George Alvarez for his support as my advisor. I thank Professors Hanspeter Pfister and Stratos Idreos for generously acting as readers for my thesis.

I thank Mel McCurrie for the many productive conversations we had, extracting poses from the datasets, and his assistance with re-implementing the EANet model. I thank Sam Anthony for his generosity in computing resources. I also thank Professor Walter Scheirer for providing feedback on a early draft of this thesis. 

Lastly, I express my warmest gratitude for my parents, for their unconditional support 

\newpage
\addcontentsline{toc}{chapter}{Abstract}





\begin{abstractpage}
Pedestrian re-identification (ReID) is the task of continuously recognising the same individual across time and camera views. Researchers of pedestrian ReID and their GPUs spend enormous energy producing novel algorithms, challenging datasets, and readily accessible tools to successfully improve results on standard metrics. Yet practitioners in biometrics, surveillance, and autonomous driving have not realized benefits that reflect these metrics. Different detections, slight occlusions, changes in perspective, and other banal perturbations render the best neural networks virtually useless. This work makes two contributions. First, we introduce the ReID community to a budding area of computer vision research in model evaluation. By adapting established principles of psychophysical evaluation from psychology, we can quantify the performance degradation  and begin research that will improve the utility of pedestrian ReID models; not just their performance on test sets. Second, we introduce NuscenesReID, a challenging new ReID dataset designed to reflect the real world autonomous vehicle conditions in which ReID algorithms are used. We show that, despite performing well on existing ReID datasets, most models are not robust to synthetic augmentations or to the more realistic NuscenesReID data.

\end{abstractpage}


\begingroup
  \hypersetup{linkbordercolor=white,linkcolor=black,
    filecolor=black, urlcolor=black}  
    
\tableofcontents


\endgroup

\newpage

\newpage
\endofprelim
\cleardoublepage

\chapter{Introduction}
\label{introduction}
Pedestrian re-identification (ReID) is the problem of determining whether two pedestrians share the same identity. In the context of computer vision, this can be across camera views and/or across time. Pedestrian ReID algorithms are essential in the development of autonomous vehicles.  They play a crucial role in keeping track of pedestrians across longer time periods by supplementing physics based tracking models when pedestrians move in and out of the vehicle's field of view \cite{chen_tracking} \cite{Ciaparrone2019DeepLI}  \cite{deepsort}. Keeping track of pedestrians across longer periods of time is important because of it's down stream use. For example, it allows for better anticipation of what a pedestrian may do, such as crossing the road, ultimately increasing the safety of autonomous vehicles.  

The pedestrian ReID community broadly conflates the terms identification and re-identification. In general, pedestrian ReID refers to the problem of correctly associating spatio-temporally separated views of the same individual, for example, across a camera network. Pedestrian identification generally refers to the problem of matching an image of a person to an existing database of identities. This work, and the majority of the literature in this area, focuses on re-identification. 


ReID has been studied for a number of years. While there has been substantial progress, ReID algorithms still fail in scenarios which are frustratingly easy for humans. ReID algorithms often struggle with changes in lighting, pose, alignment, background cluttering, distinct camera viewpoints and occlusion. While it is easy for a person to recognise another human as the same in a variety of different lighting and vantage points, this is a highly non-trivial task for a computer. 

The typical evaluation framework for ReID consists of presenting the algorithm with a query image and a set of gallery images. The algorithm attempts to return an image from the gallery set with the same id as the query. The primary measure of performance in the ReID community is the rank-1 accuracy or mean average precision (mAP) on a standard dataset. While summary statistics of performance on ReID data sets are useful to quickly compare model performance, they do not tell us how and why the model fails. In a real world scenario, it is very useful to know when an algorithm is likely to fail so that we can compensate. Furthermore, in real world applications of ReID there are often a far wider range of conditions than seen in current data sets. For example, ReID in an autonomous vehicle context must deal with all kinds of weather partially obscuring views of pedestrians, as well as occlusion from objects such as cars.

We cannot quantify an algorithm's performance in these conditions using a single summary score. Rather, to address these issues we use visual psychophysics, a technique used to study vision in neuroscience and psychology. Psychophysics is the quantitative study of the behavioural response that a controlled stimuli provokes from a subject \cite{kingdom2016psychophysics}. Inspired by previous work using psychophysics to evaluate facial recognition algorithms, we develop a framework to more rigorously evaluate ReID algorithms and to determine where they fail \cite{RichardWebster_2018_ECCV} \cite{psyphy}. Perturbations of increasing strength are applied to query images and we record the mean average precision and top-1 performance of the algorithms as we increase the perturbation strength. An example perturbation may be Gaussian noise parameterized by an increasing $\sigma$. We use this framework to evaluate several well known ReID algorithms using a range of perturbations designed to reflect real world situations including: weather, compression artifacts, occlusion and misalignment. So far as we are aware, we are the first to apply psychophysics to re-identification. 

We also find that existing ReID datasets are insufficient for understanding performance in an autonomous vehicles context. They are all collected using systems of static cameras and primarily in public places such as markets. To rectify this issue we create a new dataset. We extract pedestrian bounding boxes from the Nuscenes \cite{nuscenes2019} autonomous vehicle dataset and construct a training set and a test set composed of a query set and a gallery set. We report rank-1 and mAP performance on this dataset as well as our new psychophysics results for several algorithms. 

Our contributions are twofold. Firstly, we introduce a new evaluation methodology for ReID. Secondly, we develop are new ReID data set designed to reflect the difficulty of the ReID problem in a real world autonomous vehicles context, addressing some of the shortcomings of other available data sets. 

The remaining structure of this thesis is as follows: 
In Chapter 2 we provide background on the ReID problem, existing datasets, classical models and the several deep learning models we evaluate. Then we describe the applications of psychophysics to ReID. In Chapter 3 we discuss our methodology. In Chapter 4 we present our results and discussion of the performance of models on NuscenesReID as well as our case study on alignment. Finally, we summarise our conclusions in Chapter 5. 

\cleardoublepage
\chapter{Background and Related Work}
\label{chapter2}

\section{ReID}

 In closed set ReID there is a gallery $\mathcal{G} = \{g_i\}_{i=1}^N$ with N images which belong to N different identities. Given a probe image (also often referred to as a query image) its identity is determined by: 
 \begin{equation} \label{eq:1}
i^* = \argmax_{i \in 1,2,..., N} sim(q, g_i)
\end{equation}
where sim is some similarity function, often euclidean or cosine distance on some feature vector \cite{Zheng_reid_survey}. In practice, and in most data sets, there may be multiple images per identity in the gallery set. In general, most algorithms return a ranked list of possible matches. The majority of advances in ReID have been in improving the models used to map an image of an object to a feature vector. 

The two most commonly used statistics in the literature are rank-1 accuracy and mean average precision (mAP). Both of these statistics are calculated across a test set. Rank-1 accuracy is simply the proportion of images in the query set for which the model returns a matching identity from the gallery. mAP is useful when there are multiple ground truths, in this case multiple matching ids in the gallery set. To calculate mAP we calculate the average precision (area under the precision-recall curve) for each query and then take the mean. When there are multiple correct ids this is beneficial, because it takes into account both the precision and the recall. Since we are focusing on the use of ReID in tracking, we only care about the first returned matching image and so we primarily use rank-1 accuracy in this paper. 

In closed set ReID, query identities are assumed to exist in the gallery, but in open set ReID this is not always the case. Open set systems need also to consider the pedestrian verification problem. Such systems need to add the following additional constrain to Eq. \ref{eq:1},
 \begin{equation} \label{eq:2}
sim(q, g_i^*) > h
\end{equation}
where $g_i^*$ is the closest match in the gallery to the query and h a threshold above which we assume q is a new identity that must be added to the gallery set. In \cite{openset} Liao et al. divide the open set problem into two tasks: detection, which decides if a probe identity is in the gallery, and identification which determines the ID of an accepted probe. While open set ReID is seeing increasing research attention, there is only one small dataset \cite{openset_data} that has been adapted for this task and there is not yet a consensus in the community on an evaluation protocol \cite{openset_survey}. For these reasons open set ReID is beyond the scope of this work.

\section{ReID Datasets}

    \begin{table}[h]
    \centering
    \makebox[\linewidth]{
    \begin{tabular}{lllllll}
    Dataset & Release & \# identities & \# cameras & \# images & Label Method & Crop Size \\
    VIPeR \citep{viper}& 2007 & 632 & 2 & 1264 & hand & 128x48 \\
    CUHK01 \citep{cuhk01} & 2012 & 971 & 2 & 3884 & hand & 160x60 \\
    CUHK02 \citep{cuhk02}& 2013 & 1816 & 10 in 5 pairs & 7264 & hand & 160x60 \\
    CUHK03 \citep{cuhk03}& 2014 & 1467 & 10 in 5 pairs & 13164 & hand/DPM & vary \\
    Market1501 \citep{market1501} & 2015 & 1501 & 6 & 32217 & hand/DPM & 128x64\\
    MSMT17 \citep{msmt17} \tnote{1}  & 2018 & 4010 & 15 & 126441 & Faster RCNN & vary
    \end{tabular}
    }

    \caption{Commonly used ReID data sets}
    \label{tab:dataset_table}
    \end{table}

    Table \ref{tab:dataset_table} summarizes some commonly used ReID datasets. Just as larger data sets have spurred innovation in other areas of computer vision, for example ImageNet and object classification \citep{imagenet_cvpr09}, larger ReID data sets have spurred the development of deep learning approaches to ReID. ReID data sets are limited by the relative difficulty of annotation. ID assignment is non trivial because a pedestrian may re-enter a long time after they are first observed. This also makes collaboration between annotators difficult.  
    
    Due to their size, MSMT17\footnote{There are two versions of this dataset. V1 is no longer publicly available. V2 is the same as V1 but with faces blurred out for privacy reasons. We use V2 in this paper.}, Market1501 and CUHK03 are the most commonly used, currently available, datasets for model training and evaluation.  DukeMTMC-ReID \citep{duke}, another large ReID dataset is no longer available. 
    
    All of these data sets use fixed cameras and in general are designed for pedestrian ReID from a surveilance perspective. In section \ref{chp3:nuscenes} we introduce the Nuscenes-ReID dataset, which is an ReID datset specifically for pedestrian ReID in an autonomous vehicle context. We constructed it from the publicly available Nuscenes \cite{nuscenes2019} dataset for autonomous vehicles. The details of this dataset are described in appendix \ref{appendix:nuscenes}.
\section{ReID Models}
ReID was originally used for tracking objects across a network of cameras \citep{wang2013}. ReID models can broadly be classified into classical models (using hand crafted features),  and deep learning models. Recently, nearly all advances in ReID use deep learning. This is largely due to the availability of larger data sets, such as Market1501 and improved deep learning technology and tools such as PyTorch and TensorFlow \cite{pytorch} \cite{tensorflow2015-whitepaper}. We will briefly summarize these two approaches. 

\subsection{Classical Models}
Classical models use features that can be categorized into colour and texture. Most models would use multiple features for higher discriminative power \cite{wang2013}.

Colour features were widely used because they are less sensitive to changes in pose and view point. Colour features are generally  histograms, typically over local patches or identified regions. Histograms have been used in several different color spaces, including RGB, Lab, HSV and log-HSV. In general color based features are more sensitive to changes in lighting and may not have enough discriminative power for objects that are made up of similar colors, for example two pedestrians wearing the same colored clothes \cite{wang2013}. Texture features include: Scale-invariant feature transform \cite{sift},  Color SIFT \cite{csift},  Local Binary Patterns \cite{lbp}, Speeded-Up Robust Feature \cite{surf} and histogram of oriented gradients \cite{hog}.

Examples of classical models combining color and texture features include: Gray et. al 2008 \cite{tao2008} who use the AdaBoost algorithm to select which features to use from a feature space including colour histograms and texture filters; Mignon et. al \cite{mignon2012} who construct feature vectors from colour histograms in three color spaces (RGB,  HSV  andYCrCb) and LBP texture histograms from horizontal stripes; and Zhao et. al \cite{zhao2014} who extract a 32 dimension LAB color histogram and 128 dimension SIFT description from each local 10px by 10px region with a step size of 5px. 
\subsection{Deep Learning}
\label{background:deeplearning}
 Nearly all more recent ReID papers use deep neural networks to extract a feature vector from a pedestrian bounding box. These networks typically use convolutional layers, and many of the most recent works use a pretrained backbone such as ResNet, which is then modified and fine tuned for ReID.  Deep learning was first applied to reidentification in 2014 by Yi. et al \cite{yi} and Li. et al \cite{Li}. Both of these approaches used Siamese networks. Yi partitions input images into three overlapping horizontal strips and passes each through two convolutional layers and then a fully connected layer which fuses the outputs to produce a single feature vector. Cosine distance is used to calculate the similarity between vectors. The model in Li differs quite substantially, and in particular uses a patch matching layer which multiplies together the convolutional outputs from different horizontal stripes. 
 
\begin{figure}[H] 
\centering
\includegraphics[width=0.8\textwidth]{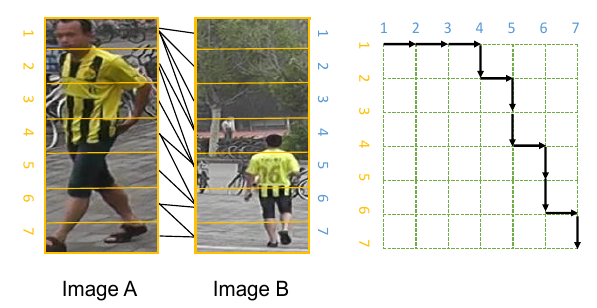}
\caption[External figure]{Example of AlignedReID local distance computed by finding the shortest path. The black arrows show the shortest path in the corresponding distance matrix on the right. The black lines show the corresponding alignment between the two images on the left. Original figure from \cite{realignedid}.} 
\label{fig:alignedreid}
\end{figure}

 More recent neural network based models are typically based on an image recognition network such as ResNet \cite{resnet}. Zhang et al (2017) introduce the AlignedReID model which attempts to solve the issue of image misalignment by automatically learning to align parts of the input without additional inputs \cite{realignedid}. The model uses two networks, one which learns a global feature and one which learns H local features for H horizontal stripes of the input. Mutual learning is used to train both networks simultaneously using both metric losses (distance between the inputs in the metric space) and classification losses (cross-entropy loss).  The global distance between the two images is the distance between the two global feature vectors. For the local distance, given local features $F = \{ f_1, \dots, f_H\}$ and $G = \{ g_1, \dots, g_H \}$, a matrix D is constructed where $d_{i,j}$ is the distance between the i-th part of the first image and the j-th part of the second image. The local distance is then the shortest path from $d_{1,1}$ to $d_{h,h}$, as illustrated in figure \ref{fig:alignedreid}.  AlignedReID uses a normalized euclidean distance and only uses the global features at inference time. Local features are used solely to improve the training of the global features. AlignedReID uses ResNet, pretrained on ImageNet, as the backbone feature extractor. 
 
\begin{figure}[H] 
\centering
\includegraphics[width=0.8\textwidth]{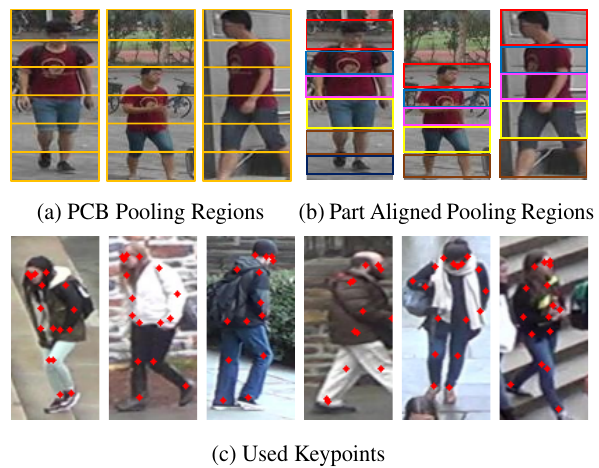}
\caption[External figure]{ (a) PCB \cite{beyondpartmodels} pools features from evenly divided
stripes. (b) PAP pools features from keypoint delimited regions (c) The
keypoints predicted by a model trained on COCO \cite{coco}, for
determining regions in (b).
Original figure from \cite{eanet}} 
\label{fig:eanet}
\end{figure}

Huang et al (2018) introduce the EANet model with the aim of improving model generalization and adaptation, noting the poor cross domain performance of existing ReID models \cite{eanet}. EANet extends on the Part-based Convolutional Baseline (PCB) model introduced in Sun et al (2017) \cite{beyondpartmodels}. In PCB, features are extracted from P horizontally segmented spatial regions, transformed using an embedding layer and then passed to a separate identity classifier for each region; the training loss is the sum of the losses for each region. At test time the concatenated feature vectors are used. Huang et al note that this is likely prone to degraded performance when misalignment occurs, see for example (a) in figure \ref{fig:eanet}. To improve part alignment,  Huang et al improve PCB by incorporating pose keypoints to refine the pooling regions, see (b) in figure \ref{fig:eanet} for the refined regions and (c) for example keypoints. The keypoints are predicted using a pose estimation network trained on the COCO dataset. In addition to the regions in (b) of figure \ref{fig:eanet}, they also include a global region and a region for the top and bottom halves of the pedestrian, to account for scenarios with poorly detected keypoints. They also explicitly handle the visibility issue, where for example a pedestrian's feet may be cut off, by using a $\Vec{0}$ feature vector for that region and ignoring it in the loss function. At test time if the p-th part of the image is not visible, the feature distance between the p-th part and all of the gallery images is ignored. EANet uses ResNet as the backbone feature extractor.

Zhou et al (2019) introduce the omni-scale network (OSNet) model \cite{osnet}. This differs from the previously discussed models in that it is an entirely new architecture that does not rely on an existing image recognition network. The authors believe that for pedestrian ReID, both local and global features are equally important, for example the color of a shirt as well as the logo on the shirt. They apply this insight to develop a network that equally emphasizes features at all scales. They argue that earlier networks which use existing object recognition networks are at a disadvantage. This is because networks like ResNet were developed for object category-level recognition tasks which are fundamentally different from instance-level recognition tasks such as ReID. The network is constructed by using multiple parallel convolutional streams with different receptive field sizes and then using a novel unified aggregation gate (AG) to merge the streams. The different streams are constructed by stacking t 3x3 convolutions, resulting in a receptive field size of $(2t+1) x (2t+1)$. It is computationally more efficient to stack multiple smaller convolutions than to perform a larger convolution. The AG dynamically weights the outputs of the different convolutional streams. The authors introduce t as a parameter for the scale of the convolution, or the number of stacked 3x3 layers in a colvolutional stream. If we let $x^t$ represent the output of the convolutions stream with t $3x3$ layers then the fused output of the AG is given by:
$$\Bar{x} = \sum_{t=1}^T G(x^t) \odot x^t$$
where $\odot$ is the Hadamard product and G is a neural network outputting a vector of the same dimension as $x^t$, implemented as a multi layer perceptron with a single hidden layer and a sigmoid activation on the output. T is the largest convolutional scale, the authors used a value of $T=4$ for their network. The mini network G allows for dynamic weighting for the different scales to change depending on the input image, rather than having fixed weights after the training phase. 

OSNet-AIN is a further development of the original OSNet model that adds instance normalization layers (IN) \cite{osnet-ain}. IN normalizes each sample using its own mean and standard deviation, in contrast to batch normalization which uses statistics computed over the whole batch. The idea behind this is that IN removes image style differences caused by the environment, lighting differences and camera setups. In general it seems that inserting IN layers early in a CNN improves generalizability. The authors of OSNet-AIN were uncomfortable with this vague heuristic and a applied neural architecture search to determine the best places to put IN layers in OSNet to create OSNet-AIN.

 \begin{figure}[H] 
\centering

     \begin{subfigure}[b]{0.48\textwidth}
         \centering
         \includegraphics[width=\textwidth]{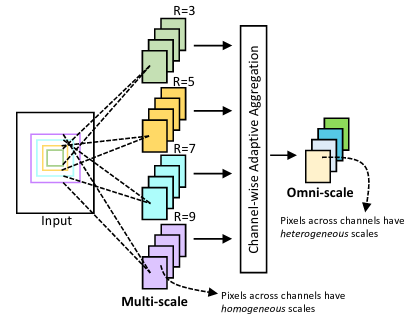}
         \caption{}
         \label{fig:osneta}
     \end{subfigure}
     \hfill
     \begin{subfigure}[b]{0.48\textwidth}
         \centering
         \includegraphics[width=\textwidth]{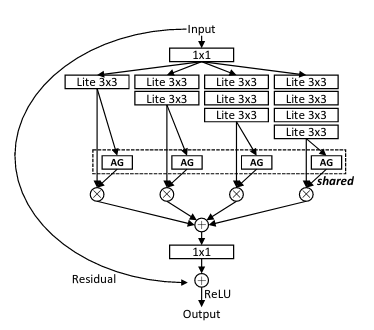}
         \caption{}
         \label{fig:osnetb}
     \end{subfigure}
        \caption{(a) Schematic of the the building block for OSNet. R: Receptive field size. (b) Bottleneck. AG: Aggregation Gate.  The first/last 1×1 layers are used to reduce/restore feature dimension
Original figure from \cite{osnet}}
\label{fig:osnet}
\end{figure}

\section{Psychophysics and its Applications to Computer Vision}
Psychophysics is the quantitative study of the behavioural response that a controlled stimuli provokes from a subject \cite{kingdom2016psychophysics}. It is a system to evaluate perceptual processes by presenting incremental perturbations of visual stimuli and recording the output. Each stimulus is varied in one or more dimensions, for example the value of $\sigma$ for Gaussian blur, which controls the difficulty of the task. We can plot the result of these experiments as an item response curve (e.g. Fig \ref{fig:translation_market}) where the performance metric on the y-axis is plotted against the manipulated dimension on the x-axis. Psychophysics is particularly useful because it allows us to probe a black box model, such as a neural network, to characterize its performance under a range of conditions. It has been used to make several discoveries in human vision; including: advancing our understanding of the human mental number line (how humans perceive numbers at different scales) \cite{dehaene2003neural}, the spectral sensitivity of the human eye \cite{bowmaker1980visual} and luminance discrimination \cite{whittle1986increments}. In all of these cases, psychophysics was applied to determine the characteristics of the human visual perception system without needing to understand the underlying biological process. It was only later that we came to understand the mechanisms behind these processes, through our understanding of cones and rods for spectral sensitivity and the neuronal structure of the brain for the mental number line. 

Psychophysics has been applied to several problems in computer vision. RichardWebster et al 2016 \cite{psyphy} develop the PysPhy frame for psychophysical analysis of object detection, using procedural graphics to perturb stimuli so as to test object recognition networks at scale. In Richard Webster et al 2018 \cite{perceptual_annotation} they extend the framework to tackle the problem of face recognition, bridging vision science and biometrics. 
Psychophysics has also been applied to reinforcement learning. Deepmind recently released the Psychlab framework \cite{psychlab}. This framework allows both human and computer agents to interact with the same tasks from visual psychophysics and be directly compared to each other. This can enrich understanding of the computer agents and contribute back to improving design. They include a range of traditional experiments in their framework
including: multi object tracking, visual search, continuous recognition and visual acuity and contrast sensitivity. Geirhos et al 2019 \cite{imagenetbias} use psychophysics to compare human and CNN performance on several experiments to show that ImageNet trained CNNs are biased towards texture. 

There are several other works in explainable AI that are similar in vein to visual psychophysics. Fong et al 2017 \cite{fong} use image perturbations to localize image regions relevant to the classification for any black box model. Wilber et al 2016 \cite{wilber} apply paramaterised perturbations to determine if it is possible to evade Facebook's automatic face detection. Hoiem et al (2012) \cite{Hoiem2012DiagnosingEI} investigate the impact of pertubations, including size, aspect ratio and occlusion,  on the error rate of the FGMR detector \cite{felzenszwalb2009object} and the VGVZ detector \cite{vedaldi2009multiple}. Neither of these detectors use neural networks. FGMR is a deformable part model and VGVZ uses a support vector machine with a learned kernel. The authors conclude that these detectors are primarily impacted by localisation errors and that they perform well with common views of objects and quite poorly on uncommon views. Karahan et al 2016 \cite{karahan2016image} perform a similar analysis on networks for face recognition, applying perturbations such as blue, noise, occlusion, color distortions and changes in color balance to three networks which were the state of the art at the time: AlexNet, VGG-Face and GoogLeNet. They found that the networks were relatively robust to color distortion and changes in color balance, but that performance degraded substantially with the application of the other pertubations. S{\'a}r{\'a}ndi et al (2018) \cite{vococclusion} apply several kinds of pertubations to the input of pose estimation networks to study their robustness. We use one of their methods, which occludes images with objects drawn from VOC2012, as one of our pertubations. 

To the best of our knowledge, psychophysics has not been used in the domain of ReID.

\cleardoublepage
\chapter{Methodology}
\label{chapter3} 
\section{Nuscenes-ReID} \label{chp3:nuscenes}
We develop a new dataset for ReID based on the Nuscenes autonomous vehicle dataset \cite{nuscenes2019}. The Nuscenes dataset consists of 1000 20 second long scenes collected in Boston and Singapore from a car with a full sensor suite. The dataset is curated to include "interesting scenes". Among other annotations, the dataset includes tracks of pedestrians. From these tracks we extract the 2d bounding boxes as seen from the cars 6 cameras. We use the tracks in the original training set for our ReID training set and from the Nuscenes validation set we construct our test set, as the Nuscenes test set does not have public annotations. We split the tracks in our test set such that the images from the first 60\% of the track are in the gallery set and the last 20\% are in the query set. We omit the 20\% in-between to ensure some level of temporal separation. For further details of the construction of the dataset see appendix \ref{appendix:nuscenes}.  See Fig \ref{fig:nuscenes_example} for examples drawn from the dataset and Fig \ref{fig:nuscenes_hist} for a distribution of the temporal gaps between gallery and query sets. We believe that the design of this dataset better represents real world scenarios for ReID used in tracking than other available ReID datasets. We use this new dataset along with Market1501 and MSMT17 in our evaluations. 

\section{Adapting Psychophysical Evaluation for ReID}

 The current paradigm in pedestrian re-identification is to calculate a metric such as Rank-1 Accuracy or Mean Average Precision\cite{Zheng_reid_survey} on a standard re-identification dataset and compare to previous works. Although convenient in its brevity, this is often too shallow and uninformative for the serious evaluation and comparison needed for critical applications in biometrics, surveillance, and autonomous driving. Additionally, it discourages innovation in areas such as alignment and post-processing where the benefits are more likely to be apparent in real world scenarios than small, homogeneous, hand-curated test sets.

    Our main contribution is to introduce the ReID community to a budding area of computer vision research in model evaluation. By adapting classic principles of psychophysical evaluation, well-known in psychology, this work quantifies the frustration of practitioners who watch models fail in situations obvious to humans  \cite{kingdom2016psychophysics}.

In closed set ReID there is a gallery $\mathcal{G} = \{g_i\}_{i=1}^N$ with N images which belong to N different identities. Given a probe image (also often referred to as a query image), its identity is determined by: $$ i^* = \argmax_{i \in 1,2,..., N} sim(q, g_i)$$
    where sim is some similarity function, often euclidean or cosine distance on some feature vector. \cite{Zheng_reid_survey} In practice, and in most data sets, there may be multiple images per identity in the gallery set.
    
    In each iteration of the M-Alternative forced-choice match-to-sample psychophysics procedure (M-AFC), a (generally human) subject is provided a sample stimulus and the response is recorded. The subject is then given a refractory period to allow their response to return to neutral. Next, the subject is provided with an alternate stimulus and subsequently given another refractory period. This is repeated M times and then the subject is forced to choose the alternate stimuli that best matched the sample. The experiment is repeated for a range of stimulus levels, created by varying the strength of the perturbation applied to the sample stimulus. We can then generate an item response curve of some summary statistic, per iteration against the stimulus levels.

    Similarly to how RichardWebster et al apply the M-AFC procedure to face recognition in \cite{RichardWebster_2018_ECCV} and to general object detection in \cite{psyphy}, we re-frame the ReID problem as a M-AFC procedure. A ReID system is provided with a probe image with unknown identity and forced to determine the identity by matching the probe against the M known identities in the gallery. In this way, the M-AFC  match-to-sample trial is equivalent to 1:N identification in biometrics. 
    
    In our framework, we perturb the probe images and then measure the resulting performance using the rank-1 accuracy. We choose to evaluate on the test set, rather than the training set, to better reflect real world conditions. 
\subsection{Psychometric Curve Fitting}

  RichardWebster et al \cite{psyphy} use area under the item-response curve (AUIRC) to summarize their item response curves. As they note in their paper, care must be taken with this statistic when applied to perturbations with unbounded parameters (e.g. $\sigma$ for Gaussian blur), as the statistic is dependent on whatever bounds the experimenter chooses.

    We instead use another technique from psychophysics, psychometric curve fitting. 
    We use this approach because we expect performance to tend towards zero as we increase the stimulus level for an unbounded perturbation and an extension of the item response curve with near zero values should not impact the function parameters for a good choice of function. We choose to fit our results to logistic functions with the form $y = \frac{c}{1 + e^{-k * (x-x_0)}} + y_0$ where the parameter of interest is $x_0$, which in this parameterization controls the midpoint of the fitted curve. In psychophysics this is referred to as the threshold. For our choice of function the threshold corresponds to the pertubuation intensity for which the model achieves half of its unperturbed accuracy. Lower threshold values tell us that an algorithm is more robust, i.e for higher stimuli levels its performance relative to its peak performance decreases at a lower rate than an algorithm that produces a curve with a higher k value. We fit the curve using least squares regression. 
     
\section{Representative models}

From the subset of researchers that have chosen to release their work, we choose a sample of representative models to analyze. We report results for: Resnet-50 with a 512 fully connected layer trained using softmax as implemented in \cite{zhou2019torchreid}, Osnet and Osnet Ain \cite{osnet}, AlignedReid++ \cite{realignedid}, the feature extractor from DeepSort \cite{deepsort} and the PAP mode of EANet\footnote{We only evaluate EANet on market 1501 since it relies on an unreleased pose estimation frame work and the poses were only published for market 1501.} \cite{eanet}.  We focus on convolutional neural network based models as classical models, such as hue saturation value histograms and histograms of oriented gradients are no longer competitive \cite{Zheng_reid_survey}. Furthermore, AlignedReid and EANet are designed to be more robust to poor alignment, but there is no analysis of this in their respective papers other than an improvement in rank-1 performance on an entire dataset. See section \ref{background:deeplearning} for further background on these models. 
\section{Targeted Pertubations}
 \begin{figure*}[h!]
    \centering
    \includegraphics[width=\textwidth]{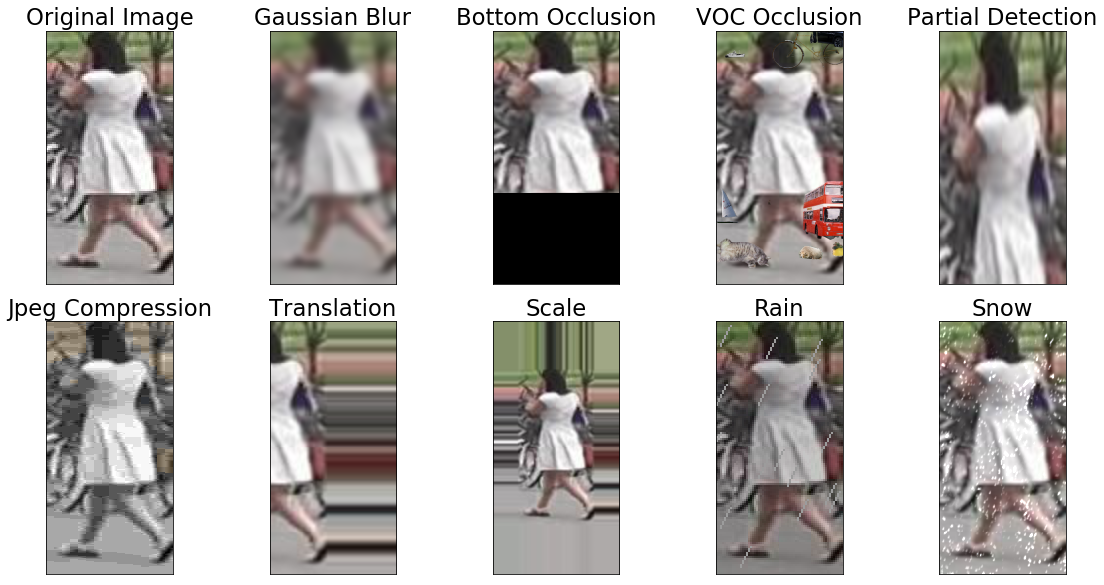}
    \caption[LoF entry]{\textbf{Example Perturbations}
    }
    \vspace{-3mm}
    \label{fig:pertubations}
    \end{figure*}
We choose perturbations that reflect real world conditions occurring in pedestrian tracking scenarios. To simulate a common bounding box problem caused by occlusions and bad detections, we crop the bounding box systematically from each side. We also occlude instead of cropping to take into account algorithms that align the pose. To simulate the image compression often used in autonomous vehicles to meet real time data transfer needs, we apply different levels of JPEG compression. Finally, we have another occlusion using natural objects drawn from the VOC20212 dataset \cite{voc2012}. We do this because we also want to see how models perform when occluded with objects that are more realistic than simply setting pixels to black.  
    
    Figure \ref{fig:pertubations} shows examples of the perturbations used in our experiments. From left to right we have: the original image, Gaussian blur paramaterized by $\sigma$, occlusion paramaterized by proportion of image occluded, VOC occlusion paramaterized by proportion of the pedestrian occluded, partial detection paramaterized by the proportion of the image excluded before reshaping, JPEG compression paramaterized by JPEG quality, translation paramaterized by the size of the translation, scale paramaterized by the scaling factor, rain paramaterized by inverse density and snow paramaterized by density.

We focus our analysis on alignment for two reasons. First, many existing algorithms are explicitly designed to be more robust to alignment without showing evidence that this is the case. Second, there are an overwhelming number of graphs when looking at all scenarios and data sets. Additional results are in appendix \ref{appendix:graphs}.

\cleardoublepage
\chapter{Results and Analysis}
\label{chapter4}

\section{Dataset Analysis}
\begin{figure}[H] 
\centering
     \begin{subfigure}[b]{0.4\textwidth}
         \includegraphics[width=1\textwidth]{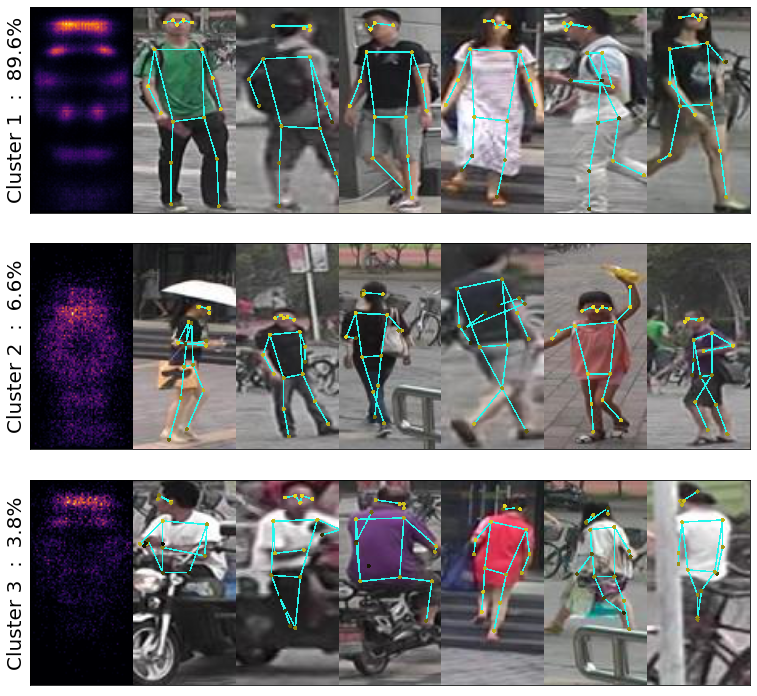}
         \caption{ Market 1501 Pose Clusters}
         \label{fig:market_alignment_clusters}
     \end{subfigure}\hfill%
     \begin{subfigure}[b]{0.4\textwidth}
         \includegraphics[width=1.0\textwidth]{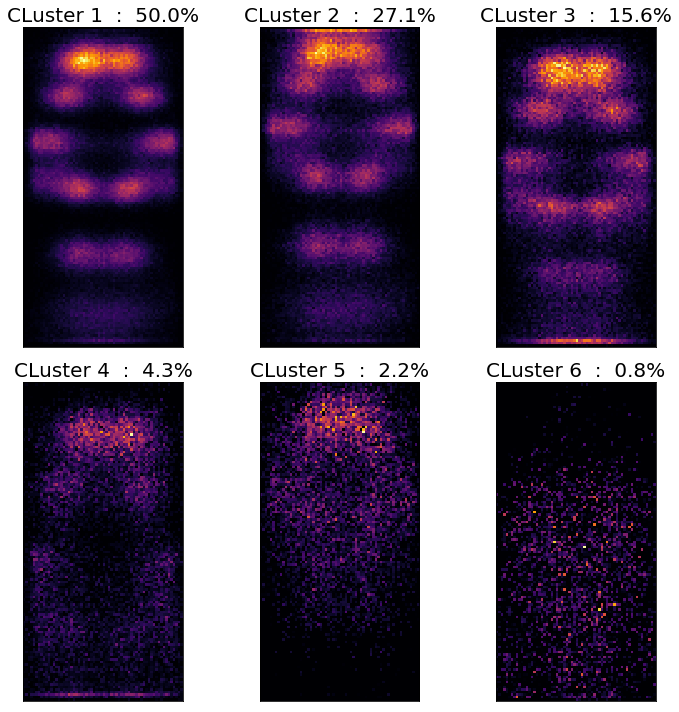}
         \caption{MSMT17 Pose Clusters}
         \label{fig:msmt_alignment_clusters}
     \end{subfigure} %
     \begin{subfigure}[b]{0.5\textwidth}
         \includegraphics[width=1.0\textwidth]{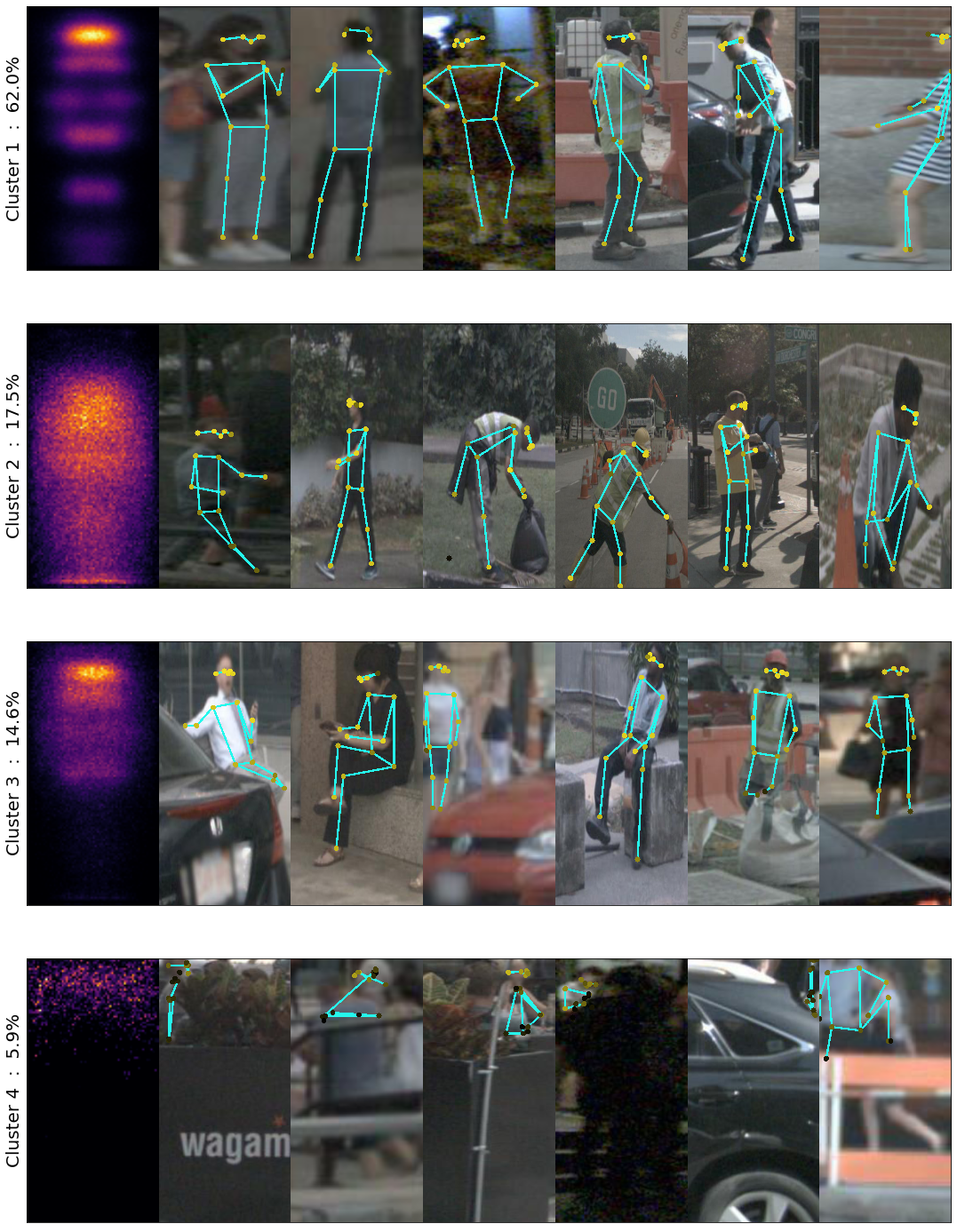}
         \caption{NuscenesReID Pose Clusters}
         \label{fig:nuscenes_alignment_clusters}
     \end{subfigure} %
        \caption{\textbf{a)} We run a state-of-the-art pose estimator over the Market1501 training set and K-Means cluster the normalized outputs \cite{pose}. The elbow method is used to select the number of clusters. Each row is a cluster. The first column is a heatmap of all joint locations from poses in that cluster, the rest are examples. 
        \textbf{b)} We perform the same analysis on MSMT17, but do not show examples due to privacy constraints. 
        \textbf{c)} We perform the same analysis on NuscenesReID with four clusters. 
        }
\label{fig:alignment_clusters}
\end{figure}

To better illustrate the differences between the datasets we extract and cluster the poses. Fig \ref{fig:alignment_clusters} depicts these clusters. On Market 1501, fig \ref{fig:market_alignment_clusters}, we find that about $90\%$ of the images fall into the first cluster, which exhibits extremely consistent major joint locations. Although less consistent, the other $10\%$ of the images in the second and third cluster are still much cleaner than poses commonly seen in applications like tracking. On MSMT17, fig \ref{fig:msmt_alignment_clusters} we find that the first two clusters contain well aligned images. Images in the third cluster are typically missing the feet. While images in the fourth cluster are typically missing the feet and knees. The fifth and sixth clusters contain people in more unusual poses, such as riding bikes or crouching down. About 77\% of the data, composed of the first two clusters, is well aligned. On NuscenesReID, fig \ref{fig:nuscenes_alignment_clusters}, we see that between 62\% and 79\% of images are well aligned, encompassing the first cluster and some images from the second. The second cluster includes people in less standard poses or who are in the lower two thirds of the frame. The third cluster contains a mix of individuals sitting down or with partially occluded lower bodies. The last cluster contains the most challenging examples, high levels of occlusion and very poor lighting. 

\section{Results on NuscenesReID}
Table \ref{tab:nuscenes_results} depicts the results for models evaluated on the same dataset they were trained on as well as on the other datasets. Osnet, which performed the best on Market-1501 with a rank 1 accuracy of 94.3 only achieves a rank 1 accuracy of 46.7 on NuscenesReID. NuscenesReID is a particularly challenging dataset because the data includes many bounding boxes with occlusions, as unlike the other datasets the bounding boxes are detected with both lidar and camera data. Furthermore, the data is collected from moving cameras in a vehicle while all other datasets use stationary cameras.

Our results also show the generally poor cross domain performance of models, which is widely commented on in the literature. Osnet AIN performs better than Osnet when trained on one dataset and tested on another. This indicates that the instance normalisation is working as intended. Additionally, models trained on NuscenesReID perform about as well, and in some cases better, on market 1501 as they do on NuscenesReID. 


\begin{table}[h!]
\begin{tabular}{|l|l|l|l|}
\hline
model (Training set)                                                                 & \begin{tabular}[c]{@{}l@{}}Market-1501 \\ R1 (mAP)\end{tabular} & \begin{tabular}[c]{@{}l@{}}MSMT17 \\ R1 (mAP)\end{tabular} & \begin{tabular}[c]{@{}l@{}}NuscenesReID \\ R1 (mAP)\end{tabular} \\ \hline
Osnet (Market-1501)                                                                  & 94.3 (82.6)                                                       & 9.5 (3.0)                                                    & 11.7 (6.8)                                                         \\ \hline
Osnet (MSMT17)                                                                       & 56.5 (28.8)                                                       & 56.8 (30.3)                                                  & 14.2 (9.0)                                                         \\ \hline
Osnet (NuscenesReID)                                                                 & 46.5 (23.2)                                                       & 7.2 (2.2)                                                    & 46.7 (36.5)                                                        \\ \hline
Osnet AIN (Market-1501)                                                              & 93.6 (80.8)                                                       & 21.7 (7.1)                                                   & 19.3 (11.8)                                                        \\ \hline
Osnet AIN (MSMT17)                                                                   & 60.1 (32.6)                                                       & 74.4 (45.6)                                                  & 17.5 (10.3)                                                        \\ \hline
Osnet AIN (NuscenesReID)                                                             & 55.0 (28.3)                                                       & 14.3 (4.5)                                                   & 48.7 (37.9)                                                        \\ \hline
\begin{tabular}[c]{@{}l@{}}DeepSort feature extractor\\  (Market-1501)\end{tabular}  & 75.3 (51.8)                                                       & 3.9 (1.1)                                                    & 9.4 (5.3)                                                          \\ \hline
\begin{tabular}[c]{@{}l@{}}DeepSort feature extractor\\  (MSMT17)\end{tabular}       & 21.9 (8.6)                                                        & 22.9 (9.0)                                                   & 6.1 (3.5)                                                          \\ \hline
\begin{tabular}[c]{@{}l@{}}DeepSort feature extractor\\  (NuscenesReID)\end{tabular} & 26.4 (10.2)                                                       & 2.1 (0.6)                                                    & 13.1 (8.3)                                                         \\ \hline
ResNet50 FC512 (Market-1501)                                                         & 90.9 (51.8)                                                       & 10.0 (3.0)                                                   & 11.5 (6.8)                                                         \\ \hline
ResNet50 FC512 (MSMT17)                                                              & 51.6 (24.6)                                                       & 66.2 (35.0)                                                  & 13.2 (7.5)                                                         \\ \hline
ResNet50 FC512 (NuscenesReID)                                                        & 46.6 (23.6)                                                       & 8.6 (2.5)                                                    & 46.3 (36.5)                                                        \\ \hline
AlignedReID (Market-1501)                                                            & 91.1 (77.7)                                                       & 6.6 (2.2)                                                    & 10.5 (6.1)                                                         \\ \hline
AlignedReID (MSMT17)                                                                 & 43.4 (21.8)                                                       & 64.2 (37.6)                                                  & 9.1 (5.6)                                                          \\ \hline
AlignedReID (NuscenesReID)                                                           & 38.9 (18.3)                                                       & 4.5 (2.6)                                                    & 39.0 (31.0)                                                        \\ \hline
\end{tabular}
\caption{Summary of model performance using summary statistics.}
\label{tab:nuscenes_results}
\end{table}

\section{Psychophysics Results}

\begin{figure}[H] 
\centering
     \begin{subfigure}[b]{1\textwidth}
         \makebox[\textwidth][c]{\includegraphics[width=0.95\textwidth]{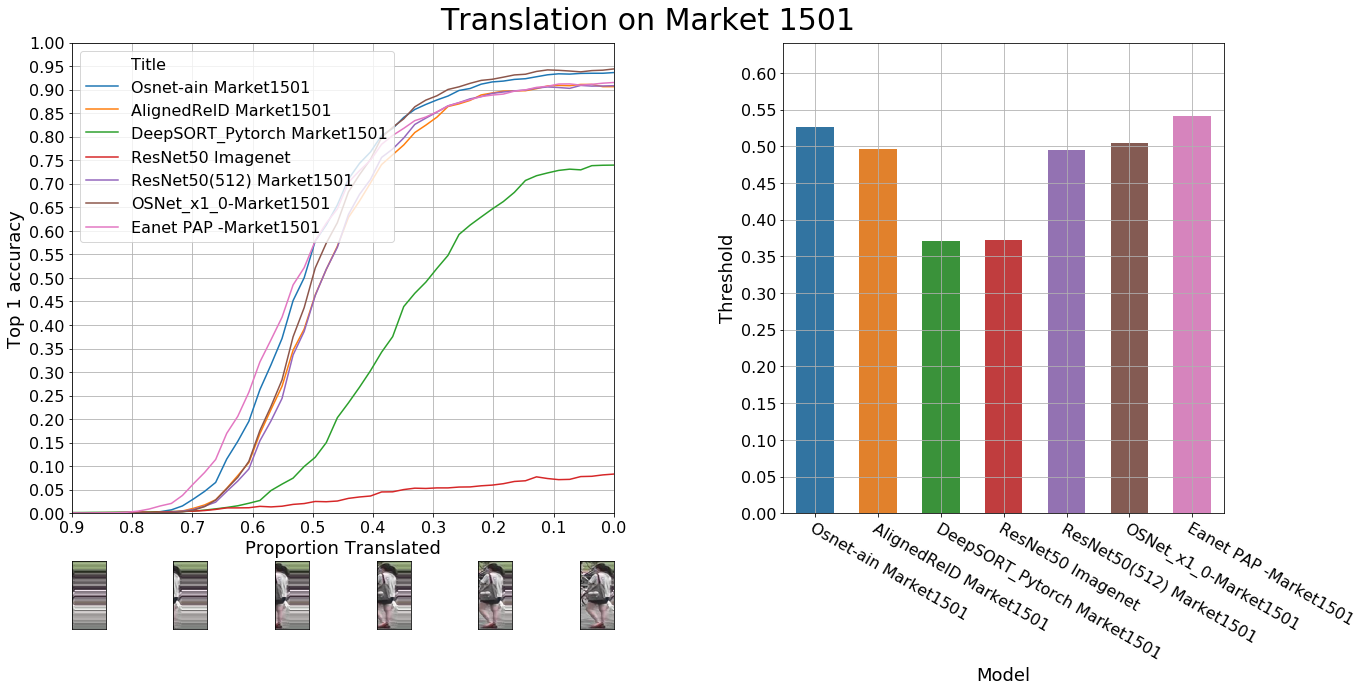}}
         \caption{}
         \label{fig:translation_market}
     \end{subfigure}\hfill%
     \begin{subfigure}[b]{1\textwidth}
         \makebox[\textwidth][c]{\includegraphics[width=0.95\textwidth]{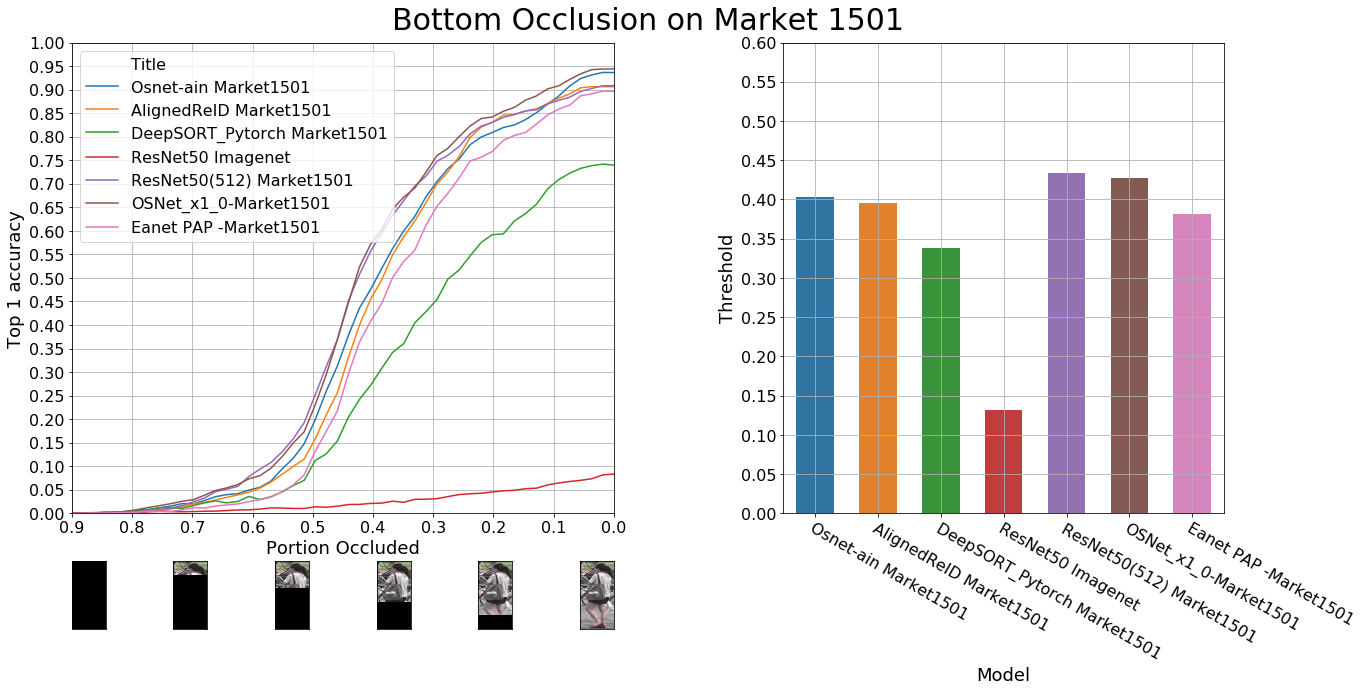}}
         \caption{}
         \label{fig:occulusion_market}
     \end{subfigure} %

     \begin{subfigure}[b]{1\textwidth}
         \makebox[\textwidth][c]{\includegraphics[width=0.95\textwidth]{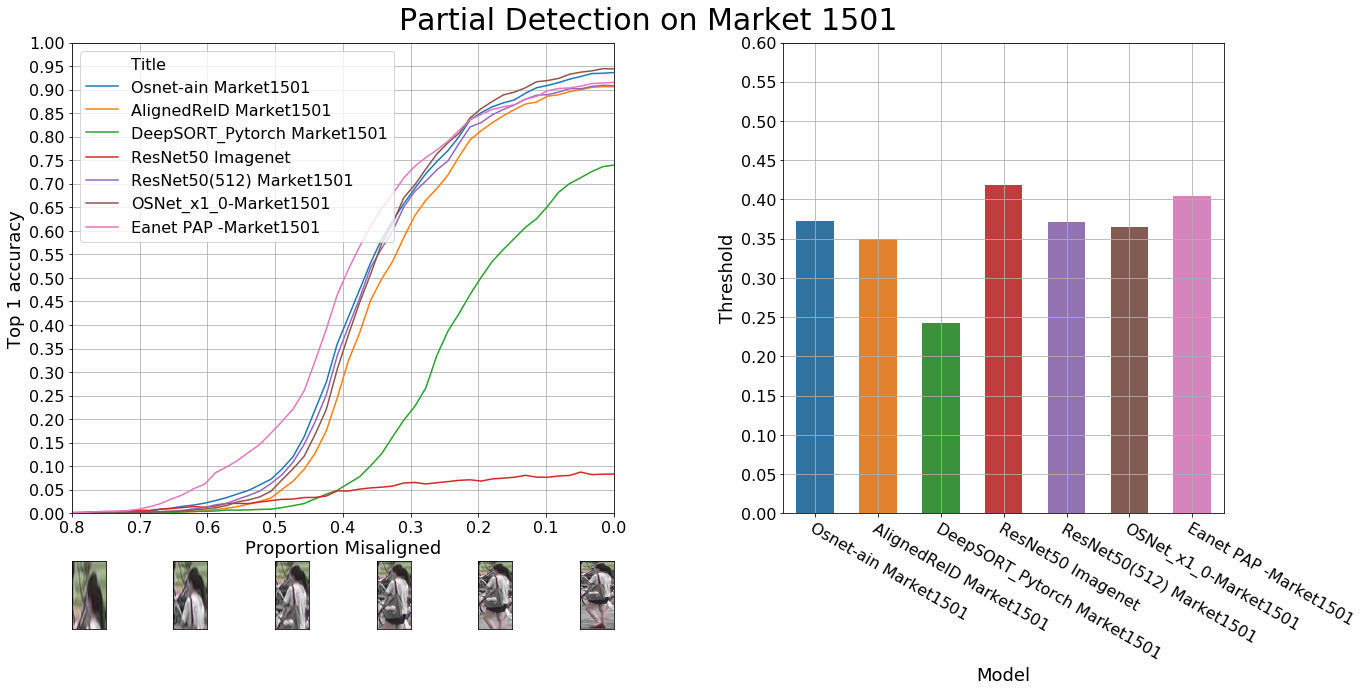}}
         \caption{}
         \label{fig:alignment_market}
     \end{subfigure}
        \caption{Psychophysics Results on Market 1501}
\label{fig:main_market_results}
\end{figure}

\begin{figure}[H] 
\centering
     \begin{subfigure}[b]{1\textwidth}
         \makebox[\textwidth][c]{\includegraphics[width=0.95\textwidth]{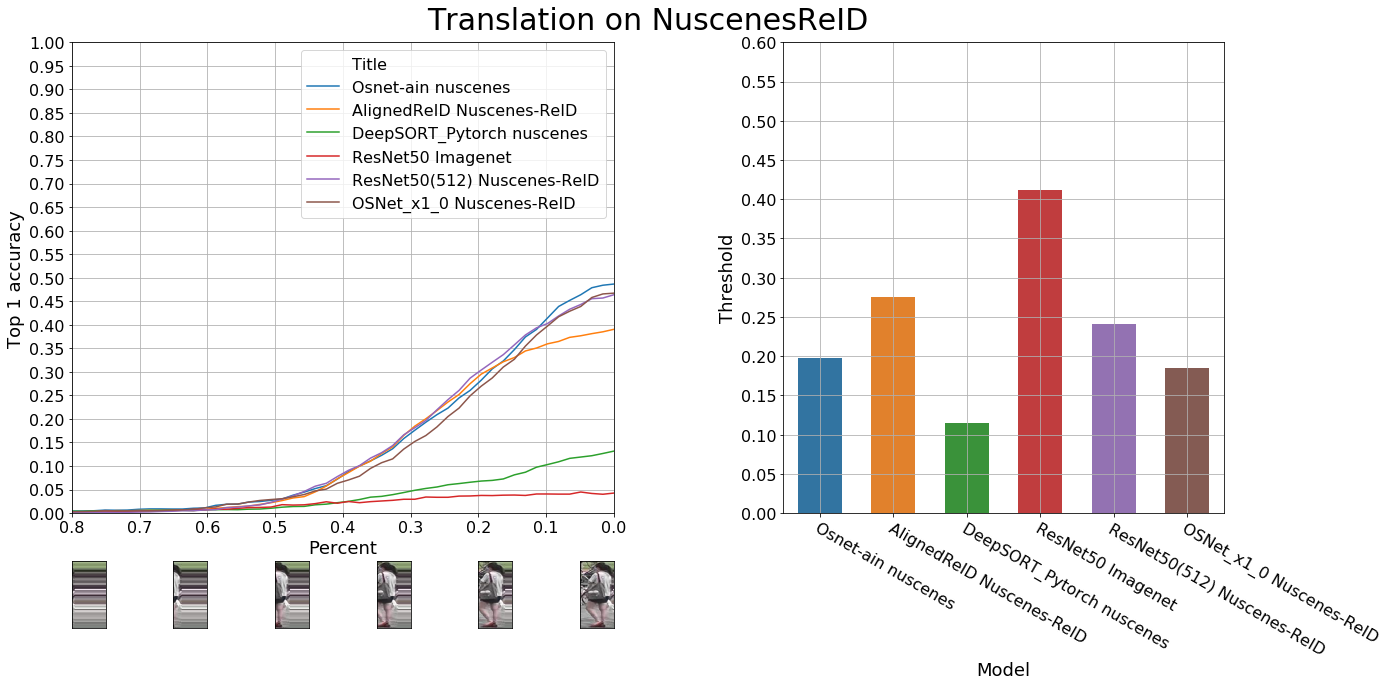}}
         \caption{}
         \label{fig:translation_nuscens}
     \end{subfigure}\hfill%
     \begin{subfigure}[b]{1\textwidth}
         \makebox[\textwidth][c]{\includegraphics[width=0.95\textwidth]{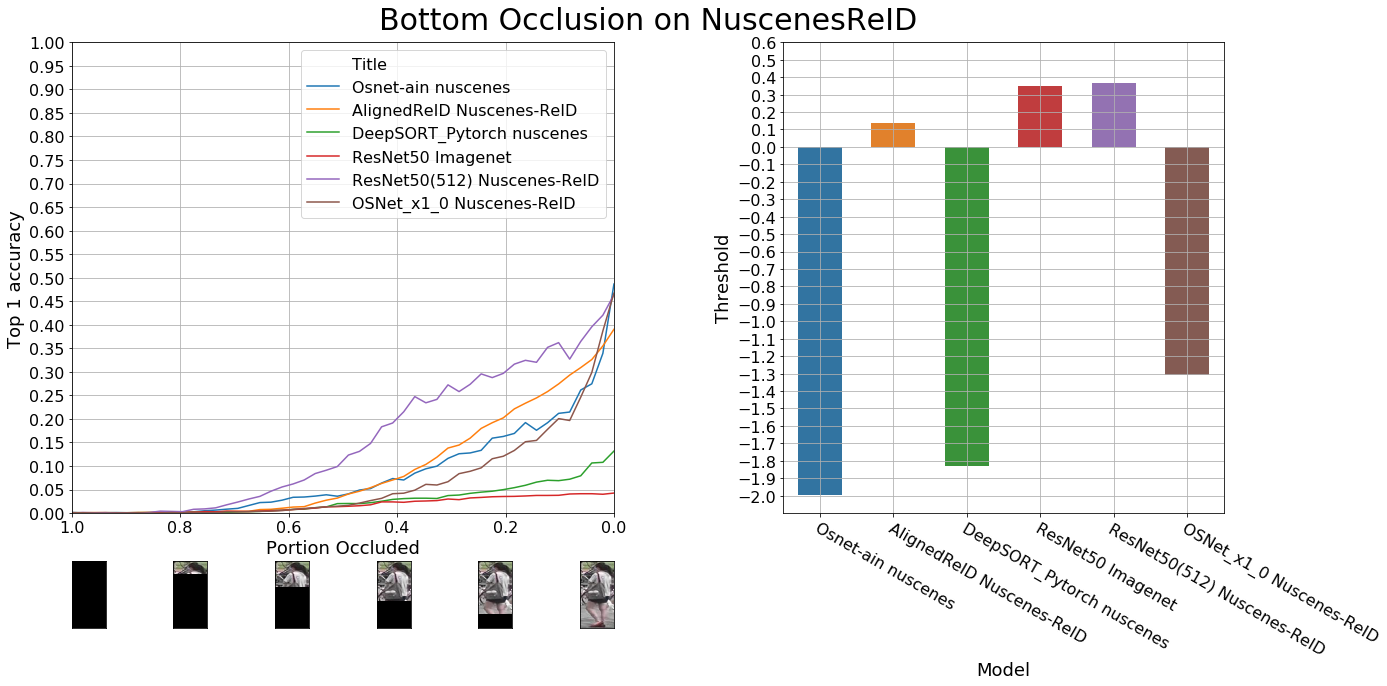}}
         \caption{}
         \label{fig:occulusion_nuscenes}
     \end{subfigure} %

     \begin{subfigure}[b]{1\textwidth}
         \makebox[\textwidth][c]{\includegraphics[width=0.95\textwidth]{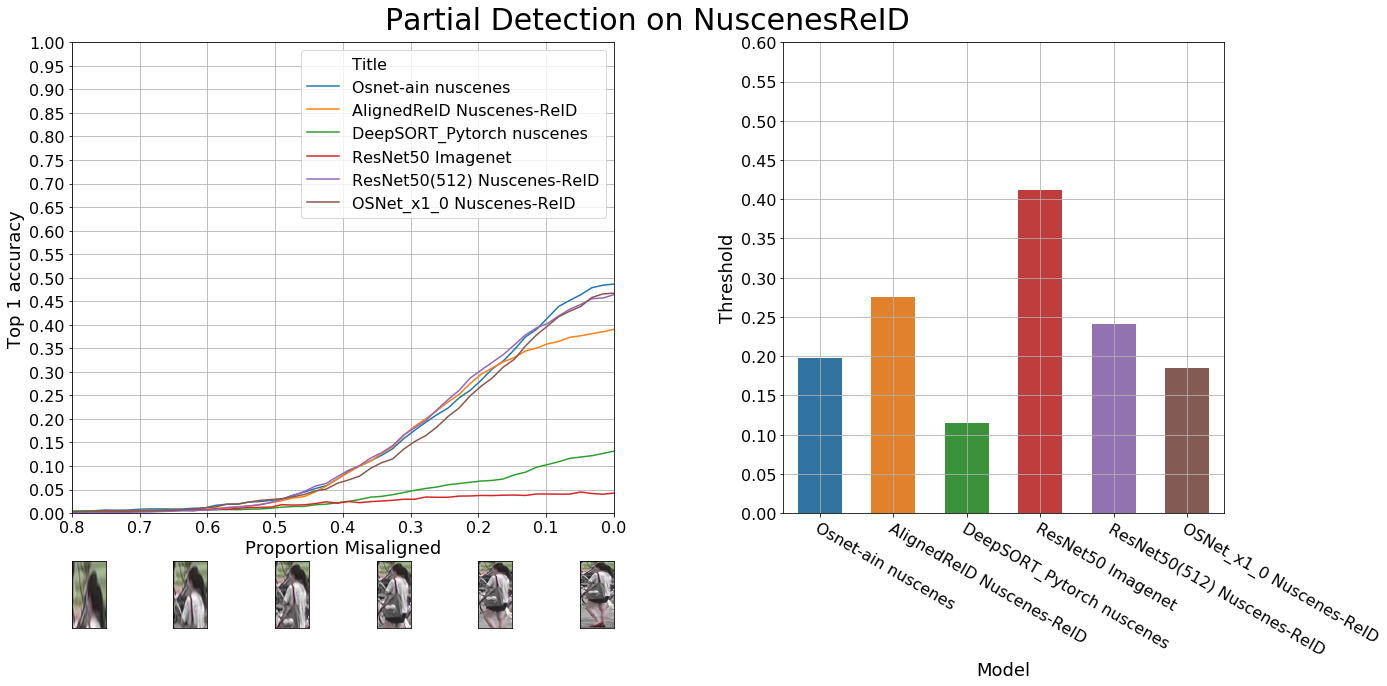}}
         \caption{}
         \label{fig:alignment_nuscenes}
     \end{subfigure}
        \caption{Psychophysics Results on NuscenesReID}
\label{fig:main_nuscenes_results}
\end{figure}

\begin{figure}[H] 
\centering
     \begin{subfigure}[b]{1\textwidth}
         \makebox[\textwidth][c]{\includegraphics[width=0.95\textwidth]{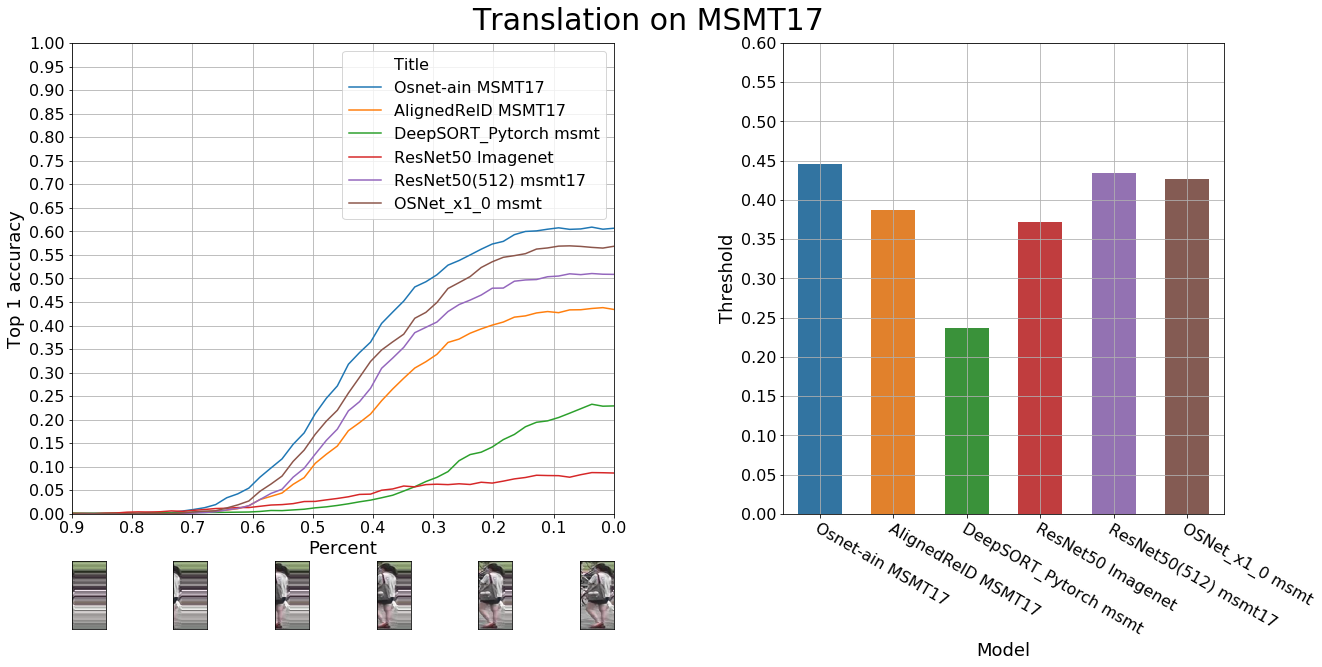}}
         \caption{}
         \label{fig:translation_msmt}
     \end{subfigure}\hfill%
     \begin{subfigure}[b]{1\textwidth}
         \makebox[\textwidth][c]{\includegraphics[width=0.95\textwidth]{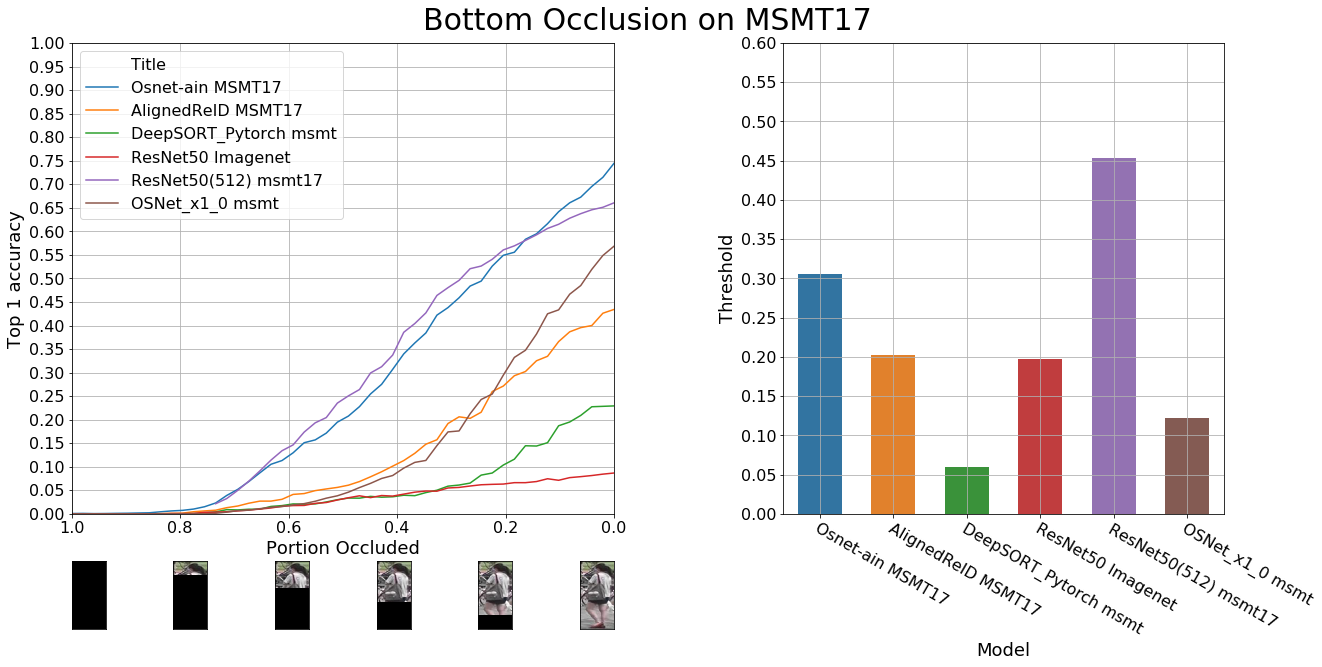}}
         \caption{}
         \label{fig:occulusion_msmt}
     \end{subfigure} %

     \begin{subfigure}[b]{1\textwidth}
         \makebox[\textwidth][c]{\includegraphics[width=0.95\textwidth]{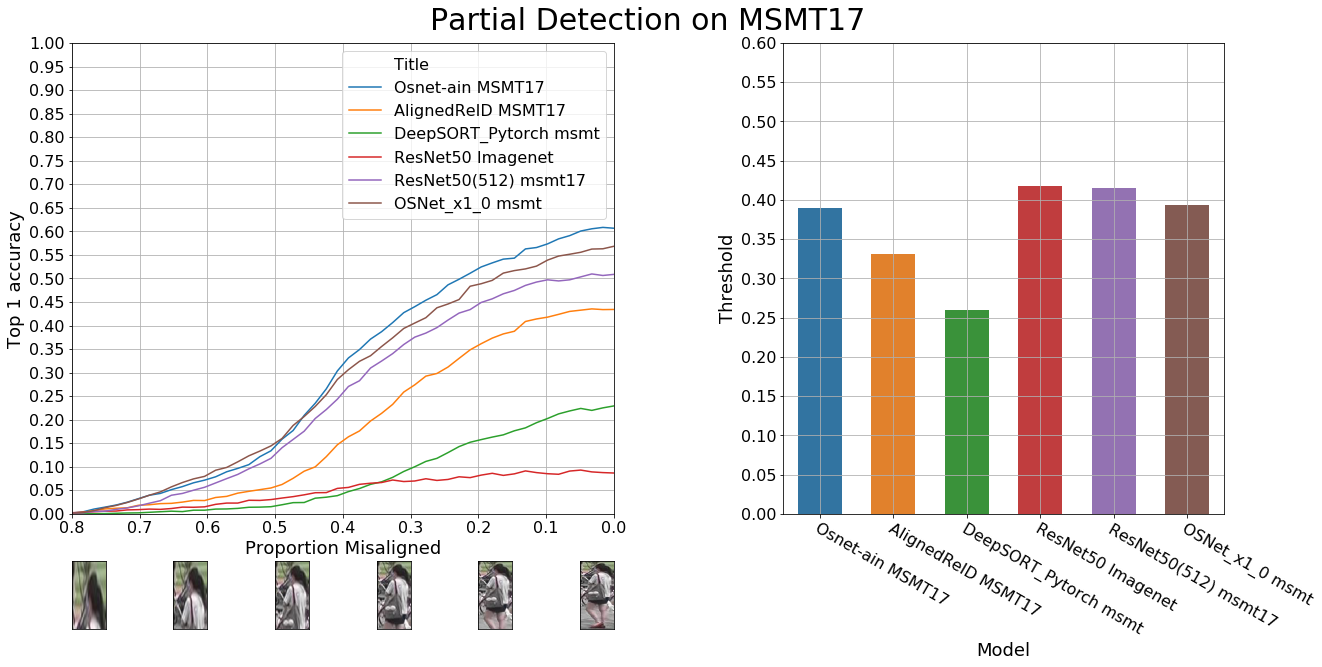}}
         \caption{}
         \label{fig:alignment_msmt}
     \end{subfigure}
        \caption{Psychophysics Results on MSMT17}
\label{fig:main_msmt_results}
\end{figure}

Generalizing across datasets, models struggle the most with occlusion, figs  \ref{fig:occulusion_market}, \ref{fig:occulusion_nuscenes} and \ref{fig:occulusion_msmt},  and perform better on partial detection, figs \ref{fig:alignment_market}, \ref{fig:alignment_nuscenes} and \ref{fig:alignment_nuscenes}, and best on translation figs \ref{fig:translation_market} \ref{fig:translation_nuscens} and \ref{fig:translation_msmt}. Models likely perform better on translation because the perturbation maintains the vertical alignment of the pedestrian and vertical alignment varies far less in the data than horizontal alignment. However, occlusion also maintains vertical alignment and models perform far worse, possibly because the black occlusion is unlike any of the training data they have seen whereas the translation and partial perturbations are much more similar to the training data. We test this hypothesis by running the same experiment, but with repeat padding instead of setting pixel values to black. Results of this experiment can be seen in appendix \ref{appendix:graphs} in Fig \ref{fig:repeat_occlusion_results}. Across all datasets, models are substantially more robust to this transformation.  This could be because the color of the last row of visible pixels is indicative of the color of the occluded pixels. 

Models may also perform worse under the bottom up occlusion because highly distinguishable features tend to be in distinct horizontal stripes of the bounding boxes. For example, shoes, pants and shirts are in distinct stripes of the image and as we occlude upwards we entirely remove the shoes and then the pants from the input. This is in contrast to a left to right occlusion which partially occludes each of the distinguishing features, eg. one shoe and half the pants. To test whether this is the case we perform an experiment occluding bounding boxes from the left to the right. The results are in appendix \ref{appendix:graphs} in Fig \ref{fig:left_occlusion_results}. Models are more robust to this mode of occlusion which supports our hypothesis.


Several of the threshold results indicate that the imagenet trained ResNet-50 is very robust. This is most evident in Fig \ref{fig:translation_nuscens} and Fig \ref{fig:alignment_nuscenes}. This is because our definition of robustness, the threshold, relates peak performance to the point where the performance degrades by half. Consequently, models that consistently perform poorly are nonetheless robust. It is thus important to assess together both the unperturbed accuracy and the threshold when evaluating a model. 

AlignedReID was designed specifically to be robust to poor alignment, see \ref{background:deeplearning} and \cite{realignedid} for more detail. However, in the partial detection experiment it is less robust than a ResNet-50 with an additional fully connected layer, only out performing it on NuscenesReID. The authors of RealignedID created a dataset from Market 1501 called Market 1501 Partial by cropping the images by 10\%-40\%. However, they do not report the results of any other models on their dataset so it is difficult to determine whether their model is more robust to misalignment than other models. Further, they only report rank-1 accuracy and mAP, making it hard to characterise model robustness. This indicates the need for more descriptive metrics. Psychophysics is one tool which fits this need.

Osnet and Osnet AIN are both relatively robust and have the best unperturbed performance. On the partial detection task on Market 1501 and MSMT17 they perform the best, both in terms of their peak performance and robustness, see Fig \ref{fig:alignment_market} and Fid \ref{fig:alignment_msmt}. 
However, on NuscenesReID their performance on the partial detection task deteriorates more quickly than both RealignedID and ResNet-50 FC, with the Osnet item response curves crossing below the AlignedID and Resnet-50 FC response curves between a misalignment proportion of 0.1 and 0.2, see Fig \ref{fig:alignment_nuscenes}. 

EANet is the most robust model on the translation and partial detection tasks, where we can see its fatter tail. It seems that the part aligned pooling technique that it uses works well on the reshaped images, but struggles to match parts when they are fully occluded. 

Our rank-1 accuracy results on MSMT for unperturbed values are lower than those reported in the literature because we are using V2 of the dataset which blurs out a region around the face of each pedestrian. Osnet reports a rank-1 accuracy of 74.9 and ReAlignedID reports 67.6 compared to our results of 56.9 and 43.8 respectively \cite{osnet} \cite{realignedid}. We used the weights which the authors made publicly available, but these trained on V1 of the data set. This implies the importance of the facial region in distinguishing identities.

Across all of the data sets, the curves for each experiment and model are very similar. However, one interesting difference is that the tails for each of the experiments on MSMT17 are fatter than on NuscenesReID and on Market 1501. It is not clear what property of the MSMT17 data causes this. 

The same trends reoccurring across data sets shows that robustness is a consistent problem in ReID and not related to the underlying data. Larger training sets may lead to better performance, but additional work needs to be done to improve model robustness. This may require developments in model training or more sophisticated post processing that is perturbation aware. 


\cleardoublepage
\chapter{Conclusion}
In this thesis we have presented NuscenesReID, a new ReID data set based on Nuscenes. This data set is based on an existing auotonomous vehicles dataset. It contrasts with existing datasets which use stationary cameras and are designed for surveillance tasks. We expect it to be useful to researchers looking to improve ReID algorithms used to supplement tracking algorithms and as a challenging new benchmark of performance. 

We have also introduced a new methodology for analyzing ReID algorithms which we expect to be of interest to the ReID community. This technique allows us to partially open the black box of ReID algorithms to determine where they excel and where they fail. We expect that this knowledge will help inform researchers in their efforts to identify cases of model failure. 

Next we performed a case study using this methodology to analyze the performance of several commonly used ReID models on misalignment, occlusion, and translation tasks. We find that one model designed to be robust, EANet, is robust to both misalignment and translation. However, the other model designed to be robust to misalignment, AlignedReID, is less robust than a ResNet-50 with a fully connect layer. 

We believe that this psychophysics based approach provides a good baseline for improved ReID metrics and analysis because it is vastly more informative than summary metrics such as mAP and rank-1 accuracy. 


\cleardoublepage

\begin{appendices}
\chapter{Nuscenes-ReID Description}
\label{appendix:nuscenes}
\begin{figure}[H] 
\centering
\includegraphics[width=0.8\textwidth]{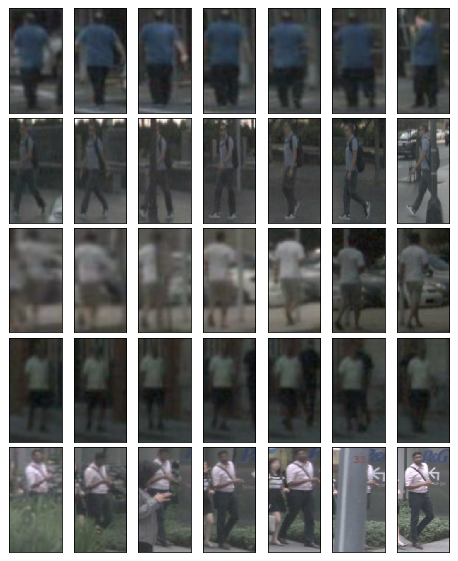}
\caption[External figure]{Example Nuscenes-ReID. Each row is a different identity.} 
\label{fig:nuscenes_example}
\end{figure}

\begin{figure}[H] 
\centering
\includegraphics[width=0.8\textwidth]{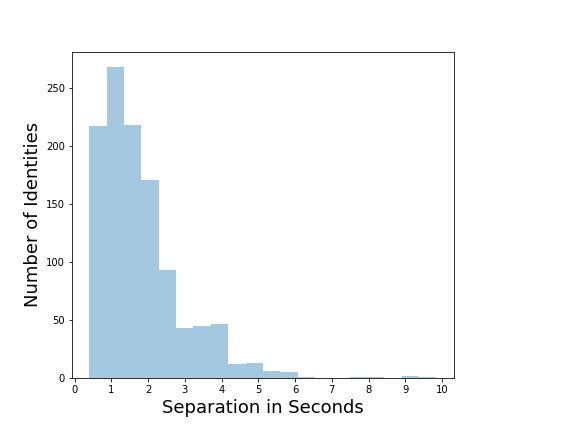}
\caption[External figure]{Distribution in temporal gaps between query images and gallery images for each id in the test set.} 
\label{fig:nuscenes_hist}
\end{figure}

The Nuscenes-ReID data-set is extracted from the Nuscenes data set. Nuscenes is a large scale autonomous vehicle data set collected in Boston and Singapore in a range of weather and daylight conditions. The data is captured from 6 cameras, 5 radars and 1 lidar,  each with a full 360 degree field of view. The original data set contains 1000 20 second long scenes with key frames annotated at 2Hz. Among other things, the dataset includes three dimensional tracks of pedestrians with bounding box annotations designed for detection and tracking research. We select all such pedestrian bounding boxes with a Nuscenes visibility score of 3 or 4 which corresponds to > 60\% of the pedestrian visible. For each pedestrian bounding box, and for each camera on which the bounding box is visible on, we project the box onto the 2d image captured by the camera and record the x and y coordinates of the top left and bottom right of the 2d bounding box. We use the Nuscenes instance token of each bounding box to determine the individual's identity. The training subset is extracted from the Nuscenes training set. The query and gallery subset are extracted from the Nuscenes validation set, since the test set does not have publicly released annotations. We select each track from the Nuscenes validation set which has 5 or more keyframes. We then put the first ~60\% of the track in the gallery set and the last 20\% in the query set. This provides a range of temporal separations between the query and gallery images of each identity. Fig \ref{fig:nuscenes_hist} shows the distribution in temporal gaps. Table \ref{tab:nuscenes-reid-table} shows basic statistics of the new dataset and Fig \ref{fig:nuscenes_example} depicts a sample of the images in the dataset. For more detail on the original data set see \citep{nuscenes2019}.

The CSV files will be publicly released and available on GitHub.

\begin{table}[h]
\begin{tabular}{l|l|l|l}
Subset  & \#ids & \#images & Mean Image Dimension       \\
all     & 8,930 & 128,019  & 75x140 \\
train   & 7,786 & 112,655  &        \\
query   & 1,144 & 4,201    &        \\
gallery & 1,144 & 11,163   &       
\end{tabular}
\caption{Nuscenes-ReID summary}
    \label{tab:nuscenes-reid-table}
\end{table}

\nobreak
\chapter{Further Psychophysics Graphs}
\nobreak
\label{appendix:graphs}
\begin{figure}[H] 
\centering
     \begin{subfigure}[b]{1\textwidth}
         \makebox[\textwidth][c]{\includegraphics[width=0.71\textwidth]{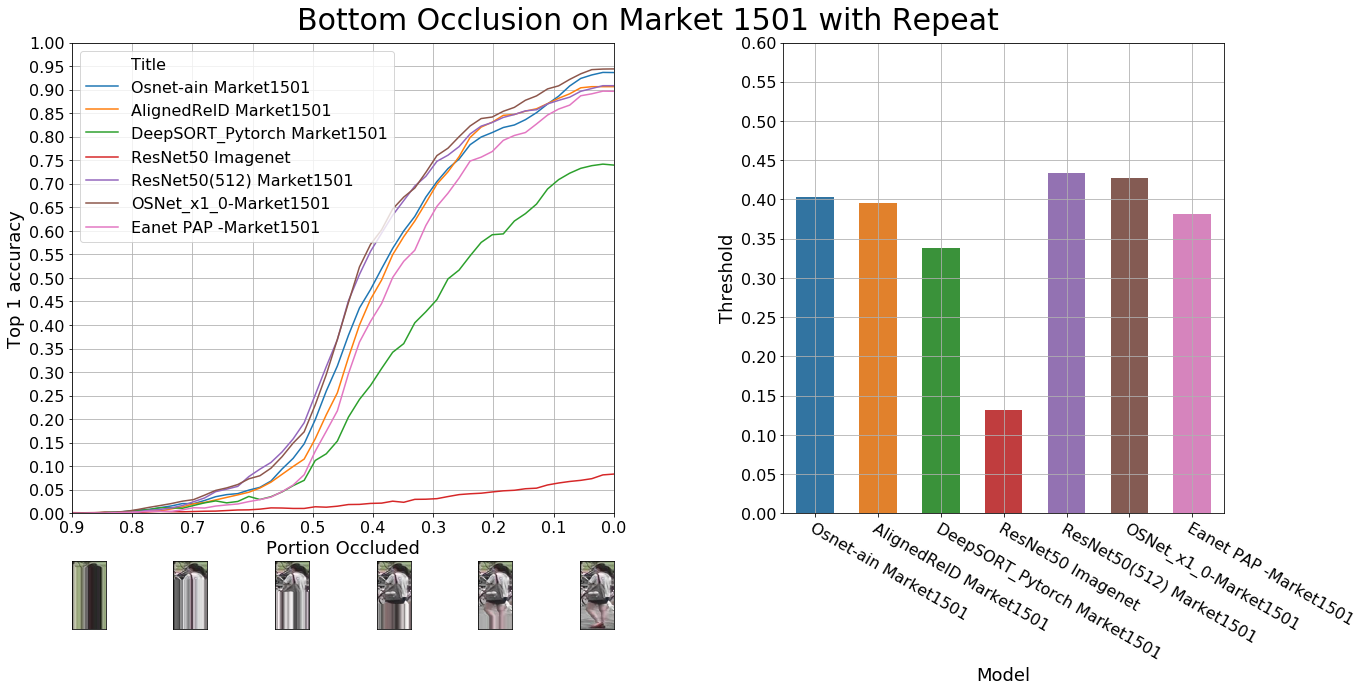}}
         \caption{}
         \label{fig:repeat_market}
     \end{subfigure}\hfill%
     \begin{subfigure}[b]{1\textwidth}
         \makebox[\textwidth][c]{\includegraphics[width=0.71\textwidth]{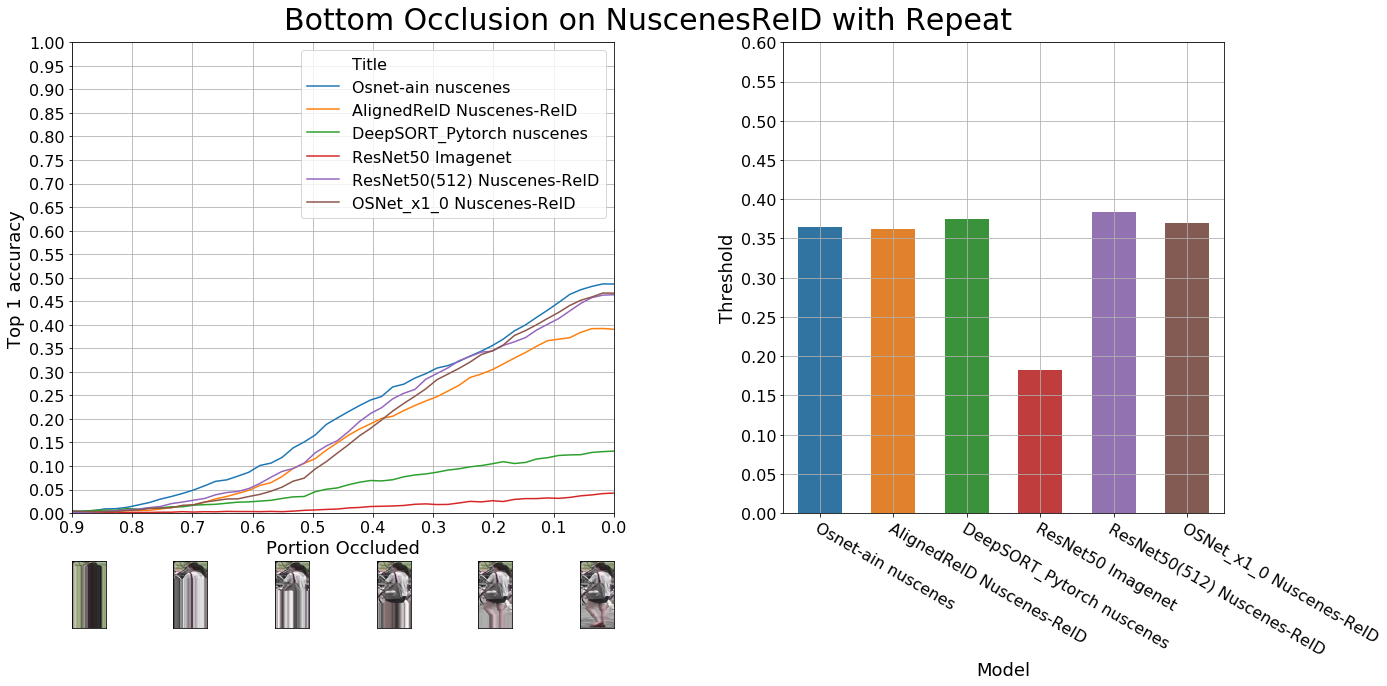}}
         \caption{}
         \label{fig:repeat_nuscenes}
     \end{subfigure} %
     \begin{subfigure}[b]{1\textwidth}
         \makebox[\textwidth][c]{\includegraphics[width=0.71\textwidth]{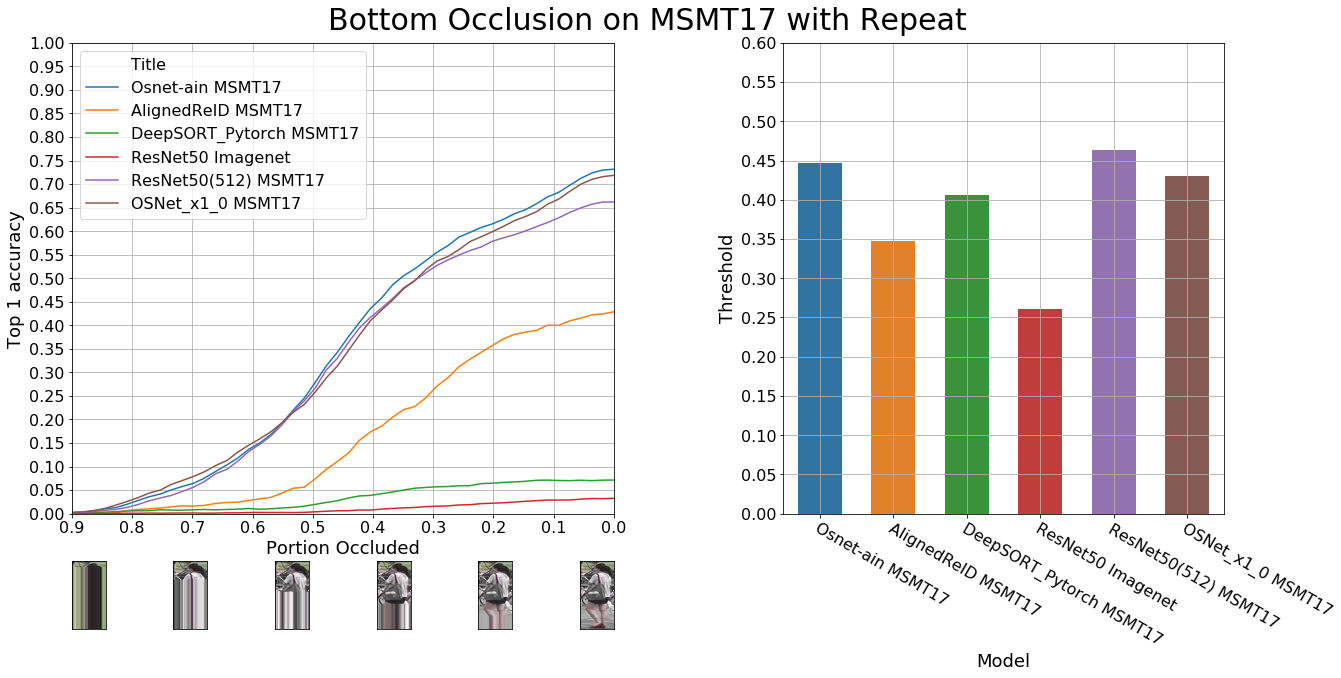}}
         \caption{}
         \label{fig:reapeat_msmt}
     \end{subfigure}
        \caption{Psychophysics Results for Repeat Occlusion}
\label{fig:repeat_occlusion_results}
\end{figure}

\begin{figure}[H] 
\centering
     \begin{subfigure}[b]{1\textwidth}
         \makebox[\textwidth][c]{\includegraphics[width=0.93\textwidth]{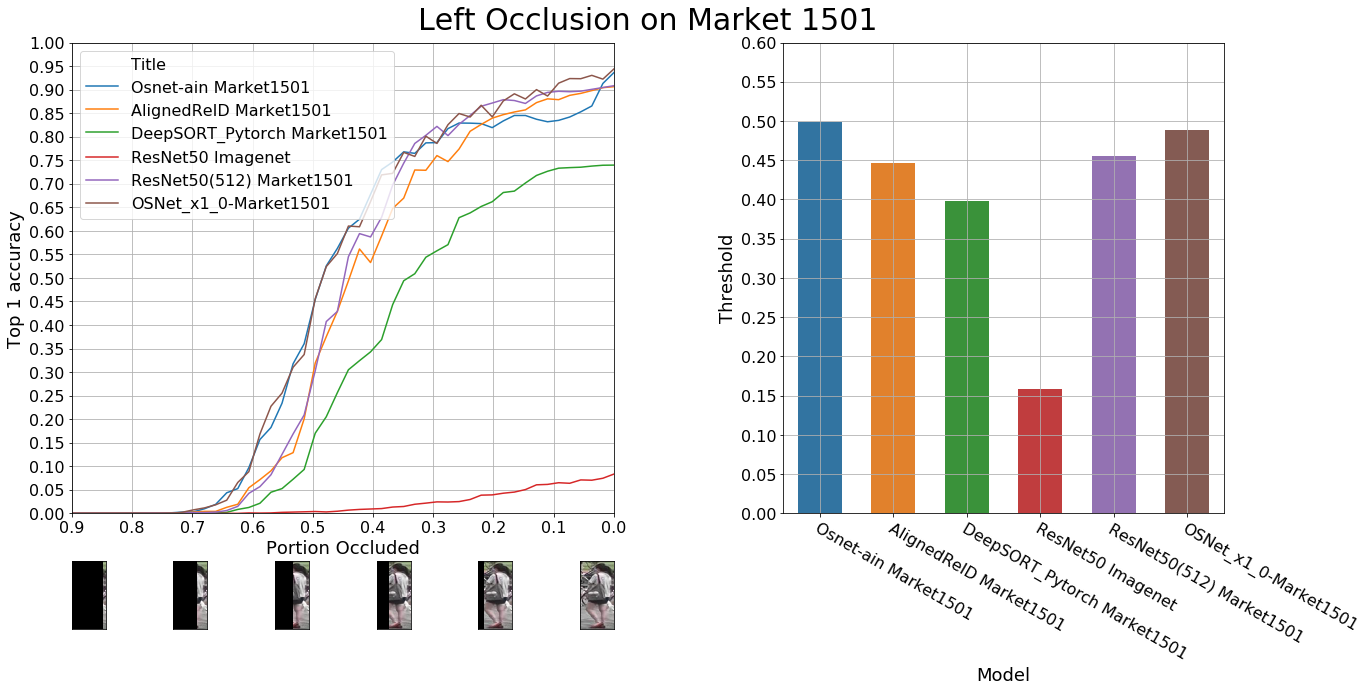}}
         \caption{}
         \label{fig:left_market}
     \end{subfigure}\hfill%
     \begin{subfigure}[b]{1\textwidth}
         \makebox[\textwidth][c]{\includegraphics[width=0.93\textwidth]{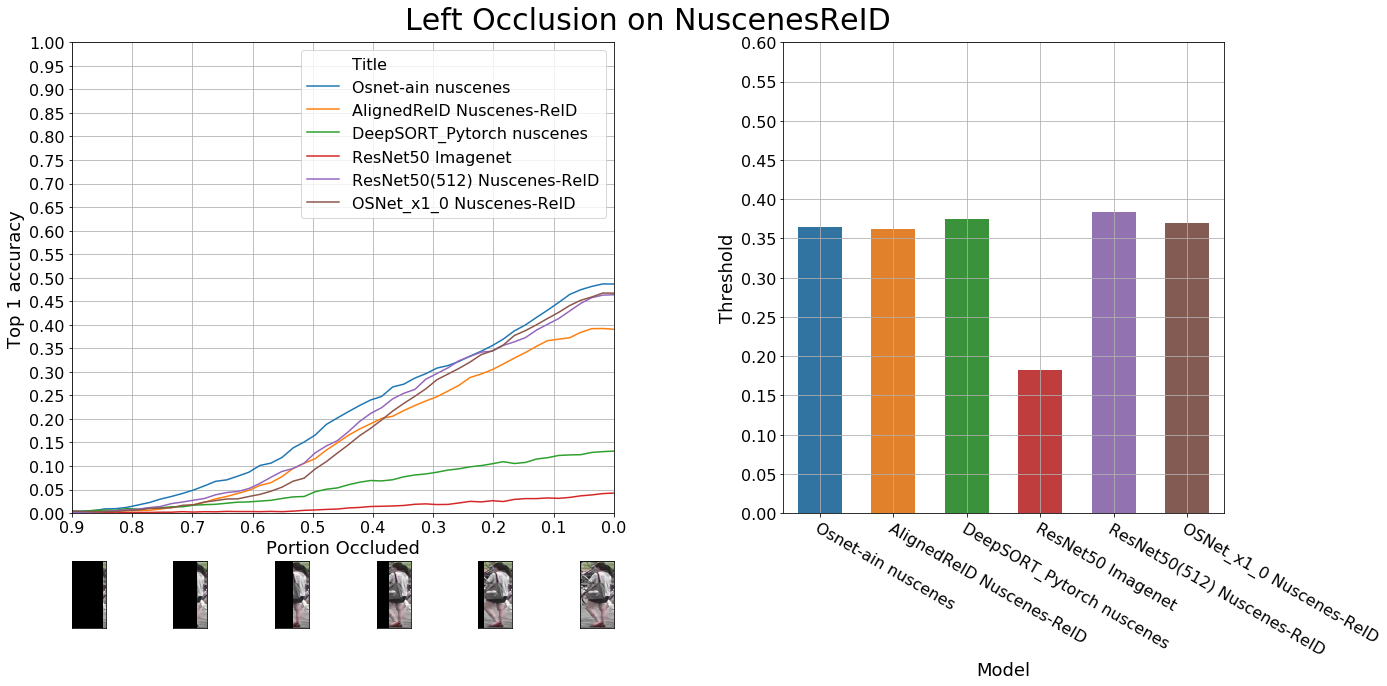}}
         \caption{}
         \label{fig:left_nuscenes}
     \end{subfigure} %
     \begin{subfigure}[b]{1\textwidth}
         \makebox[\textwidth][c]{\includegraphics[width=0.93\textwidth]{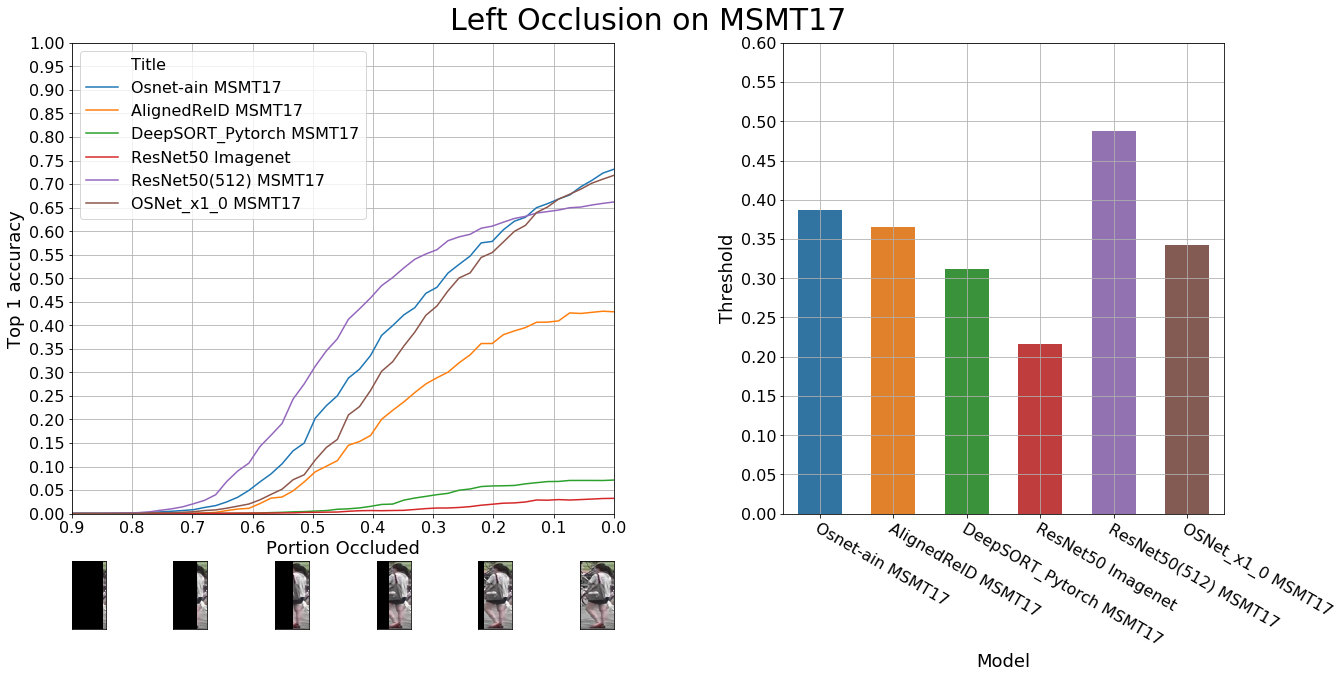}}
         \caption{}
         \label{fig:left_msmt}
     \end{subfigure}
        \caption{Psychophysics Results for Left Occlusion}
\label{fig:left_occlusion_results}
\end{figure}

\begin{figure}[H] 
\centering
     \begin{subfigure}[b]{1\textwidth}
         \makebox[\textwidth][c]{\includegraphics[width=0.93\textwidth]{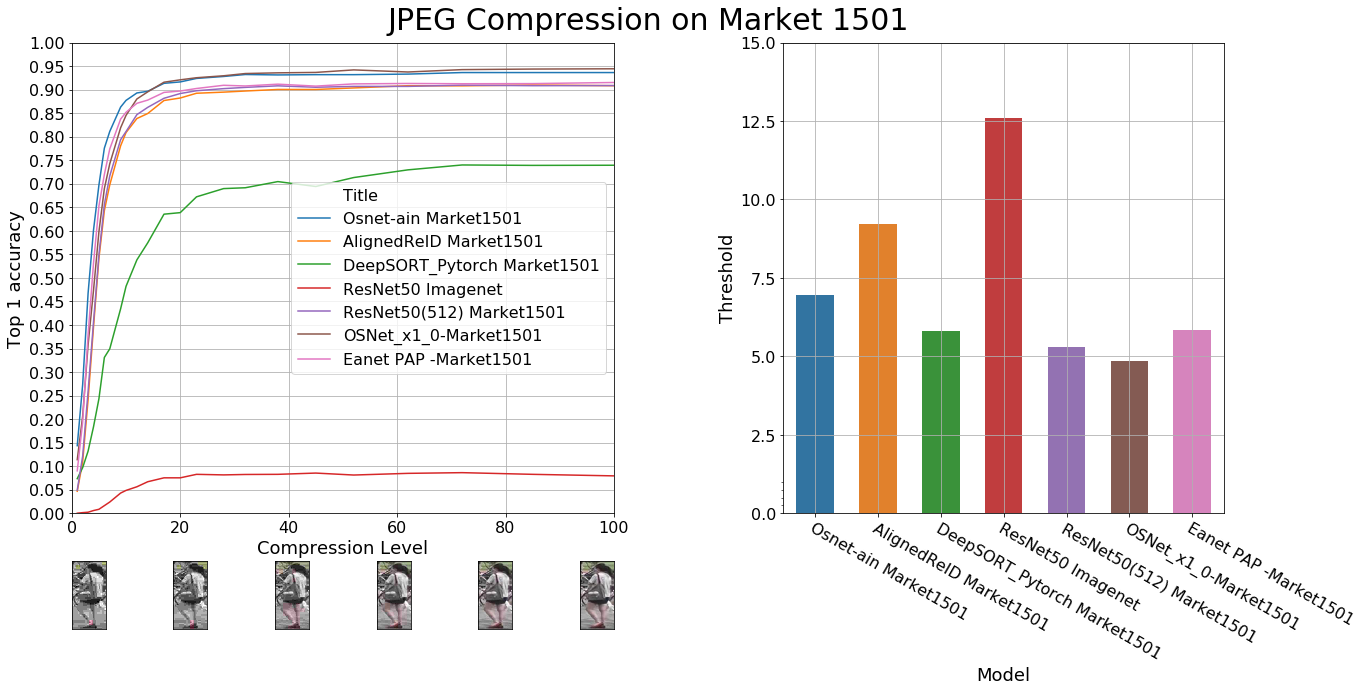}}
         \caption{}
         \label{fig:jpeg_market}
     \end{subfigure}\hfill%
     \begin{subfigure}[b]{1\textwidth}
         \makebox[\textwidth][c]{\includegraphics[width=0.93\textwidth]{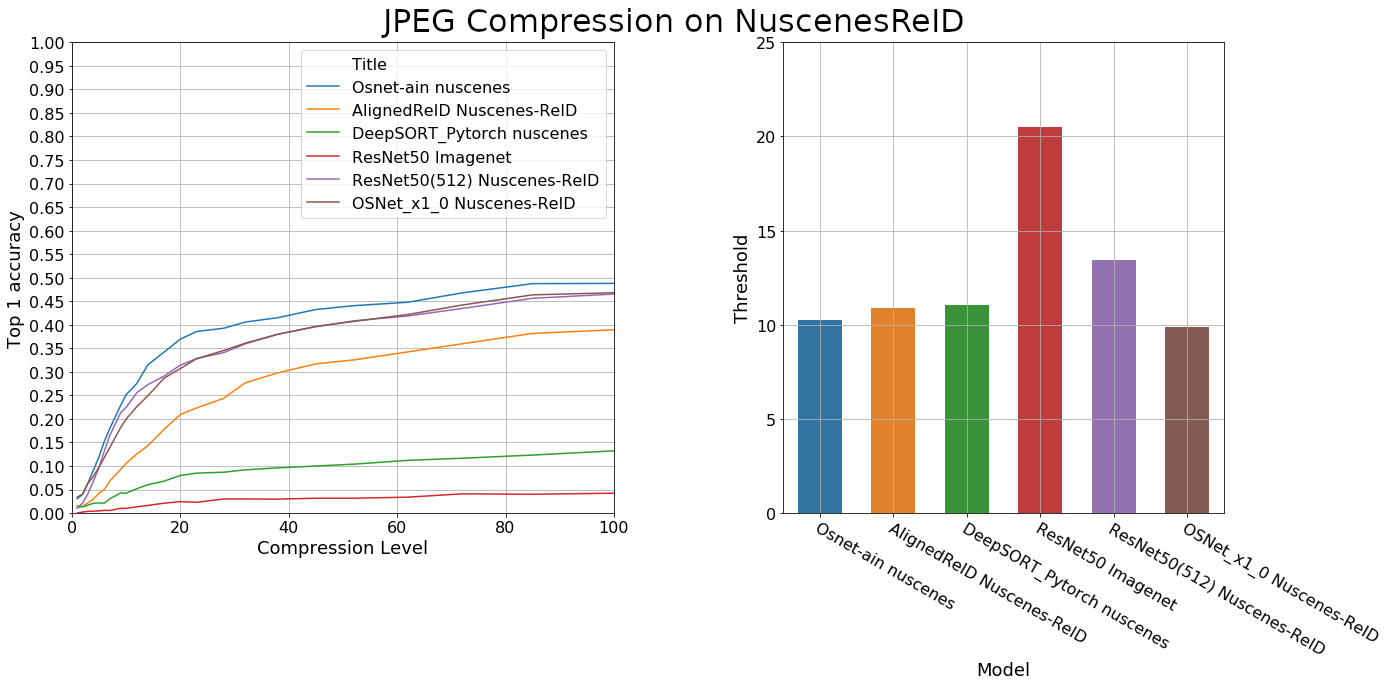}}
         \caption{}
         \label{fig:jpeg_nuscenes}
     \end{subfigure} %
     \begin{subfigure}[b]{1\textwidth}
         \makebox[\textwidth][c]{\includegraphics[width=0.93\textwidth]{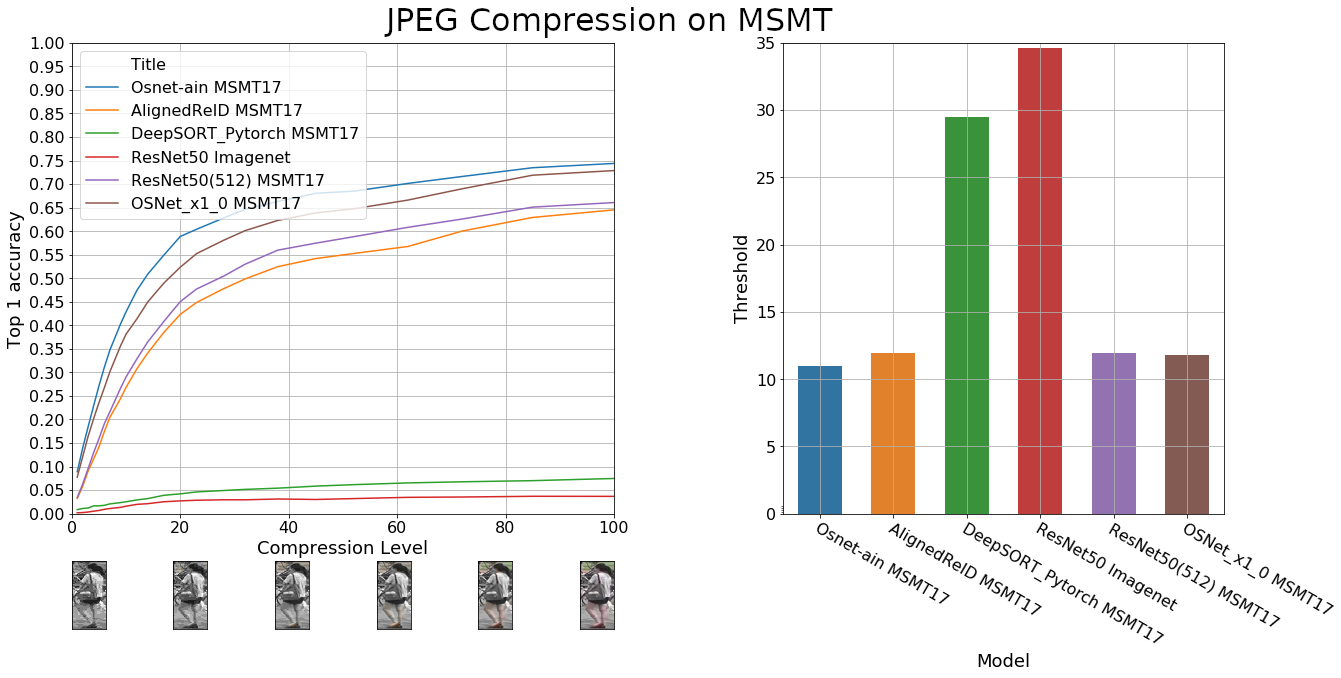}}
         \caption{}
         \label{fig:jpeg_msmt}
     \end{subfigure}
        \caption{Psychophysics Results for JPEG Compression}
\label{fig:jpeg_results}
\end{figure}

\begin{figure}[H] 
\centering
     \begin{subfigure}[b]{1\textwidth}
         \makebox[\textwidth][c]{\includegraphics[width=0.93\textwidth]{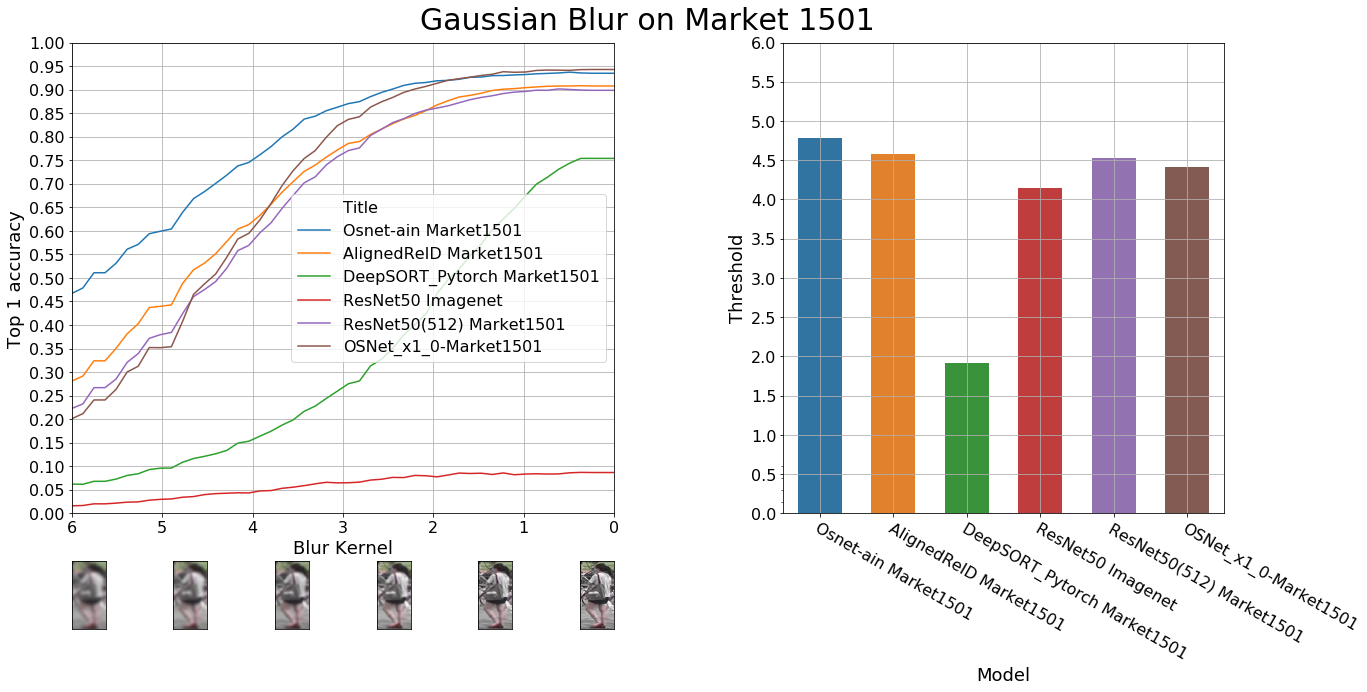}}
         \caption{}
         \label{fig:blur_market}
     \end{subfigure}\hfill%
     \begin{subfigure}[b]{1\textwidth}
         \makebox[\textwidth][c]{\includegraphics[width=0.93\textwidth]{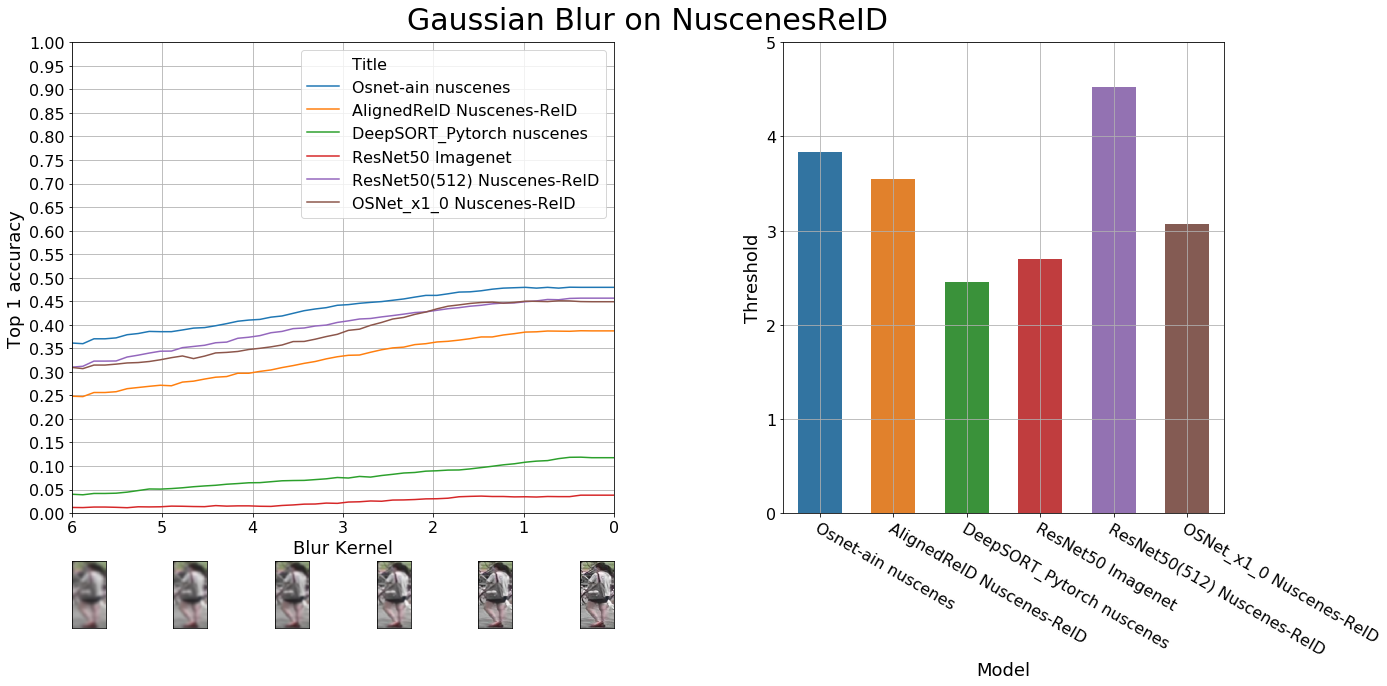}}
         \caption{}
         \label{fig:blur_nuscenes}
     \end{subfigure} %
     \begin{subfigure}[b]{1\textwidth}
         \makebox[\textwidth][c]{\includegraphics[width=0.93\textwidth]{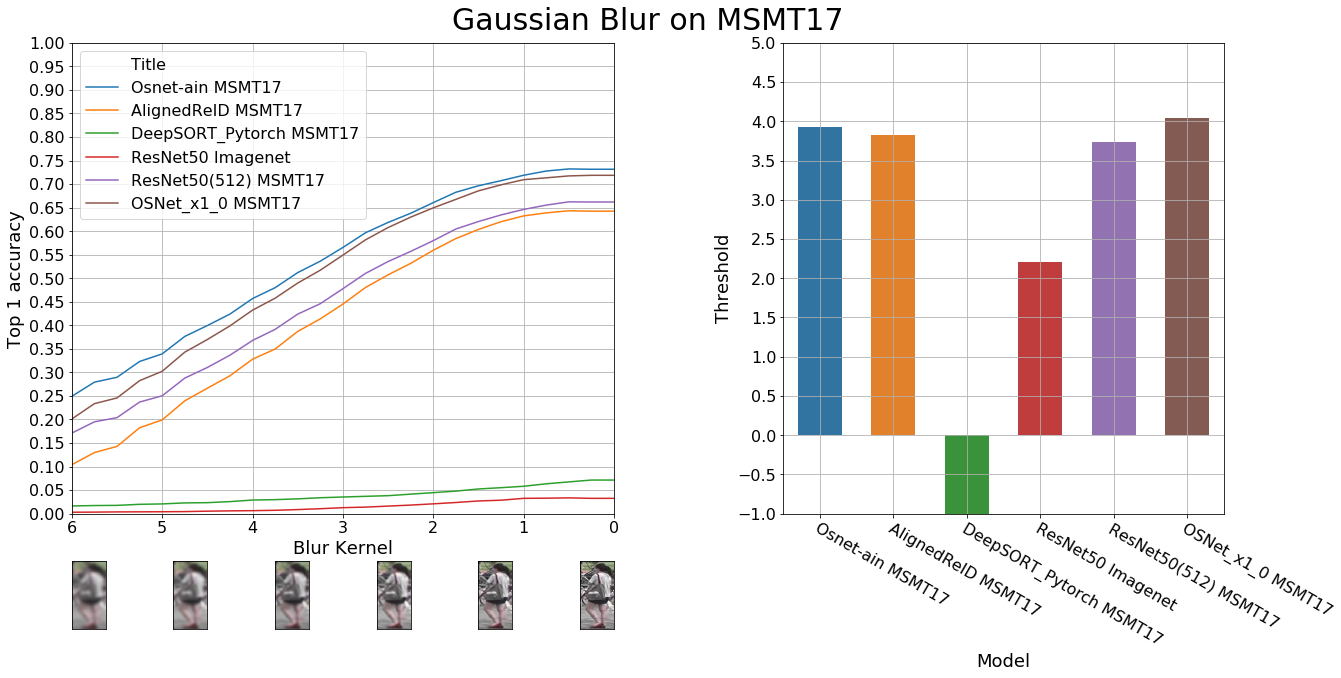}}
         \caption{}
         \label{fig:blur_msmt}
     \end{subfigure}
        \caption{Psychophysics Results for Gaussian Blur}
\label{fig:blur_results}
\end{figure}
\end{appendices}




\newpage
\singlespace
\Urlmuskip=0mu plus 1mu\relax 
\bibliographystyle{ieee}
\bibliography{Thesis.bib} 
\end{document}